\documentclass{article}

\usepackage[main,preprint]{neurips_2026}

\usepackage{microtype}
\usepackage{graphicx}
\usepackage{subcaption}
\usepackage{booktabs} 
\usepackage{hyperref}

\usepackage[american]{babel}
\usepackage{mathtools} 
\usepackage{booktabs} 

\usepackage{newfloat}
\usepackage{listings}
\usepackage{siunitx}

\usepackage[utf8]{inputenc} 
\usepackage[T1]{fontenc}    
\usepackage{hyperref}       
\usepackage{url}            
\usepackage{booktabs}       
\usepackage{amsfonts}       
\usepackage{nicefrac}       
\usepackage{microtype}      
\usepackage{xcolor}         

\usepackage{placeins}
\usepackage{multirow}
\usepackage{amssymb}
\usepackage{algorithm}
\usepackage{algorithmic}
\usepackage{verbatim}
\definecolor{light-gray}{gray}{0.95}
\usepackage{todonotes}
\usepackage{subcaption}

\usepackage{amsthm}
\usepackage{thmtools}
\usepackage{pgfplots}
\usepackage{wrapfig}
\pgfplotsset{compat=1.18}
\definecolor{cbblue}{RGB}{68,119,170}
\definecolor{cbvermillion}{RGB}{238,102,119}
\definecolor{cbgreen}{RGB}{34,136,51}

\theoremstyle{plain}
\newtheorem{theorem}{Theorem}[section]

\theoremstyle{definition}
\newtheorem{definition}[theorem]{Definition}

\theoremstyle{remark}

\newcommand{\precprec}{\prec\mathrel{\mkern-5mu}\prec}
\newcommand\algorithmicprocedure{\textbf{procedure}}
\newcommand{\algorithmicendprocedure}{\algorithmicend\ \algorithmicprocedure}
\makeatletter
\newcommand\PROCEDURE[3][default]{%
  \ALC@it
  \algorithmicprocedure\ \textsc{#2}(#3)%
  \ALC@com{#1}%
  \begin{ALC@prc}%
}
\newcommand\ENDPROCEDURE{%
  \end{ALC@prc}%
  \ifthenelse{\boolean{ALC@noend}}{}{%
    \ALC@it\algorithmicendprocedure
  }
}
\newcommand{\RETURN}{\textbf{return}}
\newenvironment{ALC@prc}{\begin{ALC@g}}{\end{ALC@g}}

\title{Hierarchical Conformal Classification}

\author{%
  Floris den Hengst \\
  Vrije Universiteit Amsterdam \\
  \texttt{f.den.hengst@vu.nl}
  \And
  Inès Blin \\
  Sony CSL, Paris, France \\
  \texttt{i.blin@vu.nl}
  \And
  Majid Mohammadi \\
  Leiden University Medical Center \\
  \texttt{m.mohammadi@lumc.nl}
  \And
  Syed Ihtesham Shah \\
  Vrije Universiteit Amsterdam \\
  \texttt{s.i.h.shah@vu.nl}
  \And
  Taranah Younesian \\
  Vrije Universiteit Amsterdam \\
  \texttt{t.younesian@vu.nl}
}

\begin{document}

\maketitle




\begin{abstract}
Conformal prediction provides reliable uncertainty quantification by generating prediction sets with finite-sample coverage guarantees, yet it typically ignores hierarchical relationships between classes. We introduce Hierarchical Conformal Classification (HCC), a framework that integrates class hierarchies into the structure and semantics of prediction sets. HCC uses a constrained optimization problem formulation to produce prediction sets composed of nodes at various hierarchy levels while maintaining rigorous coverage guarantees. To ensure computational efficiency, we prove that a restricted subset of well-structured candidate solutions is sufficient to maintain coverage and approximate optimality. Empirical evaluations across audio, image, and text benchmarks, alongside a preliminary user study, demonstrate that HCC outperforms state-of-the-art methods and that hierarchical prediction sets align better with human preferences.

\end{abstract}

\section{Introduction}
\begin{wrapfigure}{r}{.5\textwidth}
\centering
\vspace{-1.2em}
\includegraphics[width=.5\columnwidth]{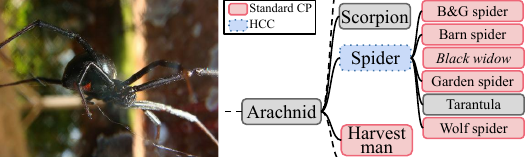}
\caption{Hierarchical prediction example from ImageNet, with standard CP output in red and HCC in blue, both covering ground truth \emph{Black Widow}. HCC yields a more compact, semantically meaningful set and was unanimously preferred.}
\label{fig:initial_analysis}
\vspace{-1em}
\end{wrapfigure}
Conformal prediction (CP) is a powerful and broadly applicable framework for uncertainty quantification, providing finite-sample statistical guarantees for any predictive model that outputs scores~\citep{vovk2005algorithmic,angelopoulos2023conformal}. Given a test input and a user-specified confidence level, CP returns a prediction set that is guaranteed to contain the ground-truth label with high probability, a property known as \emph{coverage}. Building on its strong theoretical foundations and growing adoption in high-stakes domains, recent work has extended CP to satisfy additional desiderata beyond marginal coverage, including coverage guarantees conditional on the input~\citep{angelopoulos2020uncertainty}, risk control under heterogeneous misclassification costs~\citep{angelopoulosconformal}, and the construction of structured prediction sets that respect task-specific constraints~\citep{wieslander2020deep,zhang2024conformal}.

However, most conformal prediction methods treat class labels as flat and unstructured, paying limited attention to incorporating prior domain knowledge into the construction of prediction sets with coverage guarantees. In many classification problems, domain semantics are naturally encoded through taxonomies or hierarchical relationships among labels, commonly modeled as directed acyclic graphs (DAGs). Such hierarchies arise across a wide range of applications, including disease classification in medical diagnosis, fraud type categorization in financial systems~\citep{bennett2013financial}, and topic organization in text classification~\citep{zesch2007analysis}. A prominent example is ImageNet~\citep{russakovsky2015imagenet}, with labels derived from the WordNet lexical database~\citep{miller1995wordnet}. 

\begin{figure*}[t]
    \centering
    \includegraphics[width=1\textwidth]{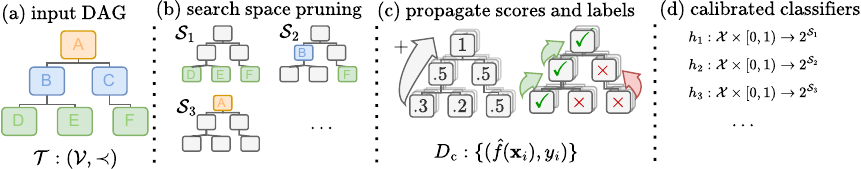}
    \caption{HCC training: (a) for an input DAG with nodes $\mathcal{V}$, (b) a set \(\{\mathcal{S} \subset \mathcal{V}\}\) of subsets that each cover all leaf nodes (the \emph{leaf covers}) is derived. (c) Classifier scores $\hat{f}(\mathbf{x}_i)$ and labels $y_i$ in a calibration set $D_c$ are propagated up the hierarchy, and (d) used to calibrate a set of conformal predictors, one for each leaf cover.}
    \label{fig:hcp_training}
\end{figure*}
\begin{figure*}[t]
    \centering
    \includegraphics[width=\textwidth]{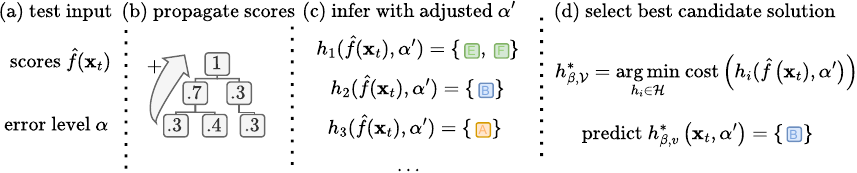}
    \caption{HCC inference: (a) the prediction scores for a test input $\mathbf{x}_t$, and for a desired error level $\alpha$, (b) classifier scores are propagated, and (c) used to infer prediction sets with the predictors $h$ obtained during training. Finally, (d) the prediction set that minimizes the cost in~\eqref{eq:objective} is selected.}
    \label{fig:hcp_inference}
\end{figure*}

Figure~\ref{fig:initial_analysis} shows an ImageNet input, with the relevant section of the WordNet lexical hierarchy along with standard and hierarchy-aware conformal classification (HCC) outputs. The inclusion of the hierarchical structure enables a prediction set that replaces a collection of semantically similar classes predicted by standard CP with a higher-level concept, yielding a prediction that is more compact, more consistent, and better aligned with the end users' understanding of the domain. 
Although some recent works have incorporated hierarchical label structure into CP, these suffer from limitations we discuss here. \citet{mortier2022set,mortier2025conformal} focus on identifying regions of high probability mass under hierarchical and representational constraints, but their methods are restricted to tree-structured hierarchies rather than DAGs. \citet{angelopoulosconformal} instead control a notion of hierarchical risk by bounding the distance between the prediction and the true label rather than focusing on leaf level coverage. Closest to our setting, \citet{zhang2024conformal} propose a learn-then-test framework that yields coverage guarantees for structured outputs in both regression and classification for DAGs. Nevertheless, their approach requires a user-specified bound on the nominal prediction set size $M$, relies on a set of predefined thresholds that should contain the optimal threshold, and involves an integer programming formulation whose scalability becomes a concern for large or deep taxonomies. 
Taken together, these methods highlight the promise of hierarchy-aware CP, but leave open the problem of constructing scalable, prediction sets with leaf-level coverage guarantees over DAGs.


We propose \emph{Hierarchical Conformal Classification (HCC)}, a principled extension of conformal classification that incorporates hierarchical label structure directly into the construction of prediction sets while preserving finite-sample coverage guarantees. Unlike existing approaches that either treat labels as flat or impose restrictive assumptions on the hierarchy, HCC operates on general class taxonomies modeled as DAGs and provides coverage at the leaf level. Central to our approach is a constrained optimization formulation that explicitly trades off prediction set compactness and semantic specificity: among all hierarchy-consistent subsets that satisfy the desired coverage guarantee, HCC selects the set that is smallest in size while remaining as informative as possible. By allowing prediction sets to contain nodes at multiple levels of the hierarchy, including both fine-grained leaf classes and more abstract ancestor concepts, HCC produces predictions that are compact, interpretable, and aligned with domain semantics.


Our solution combines learning with search. Our search step is efficient and scalable as it is conducted over a carefully constructed subspace of candidate solutions. We show that, despite the combinatorial nature of search over the whole hierarchy, it suffices to search within this much smaller, well-structured subset of candidates. 

The contributions of this paper are as follows:
\begin{enumerate}
    \item we formalize \textit{Hierarchical Conformal Classification (HCC)} as a constrained optimization problem,
    \item we propose efficient calibration and inference algorithms for HCC, including a search-based procedure that exploits the structure of the hierarchy to solve the optimization problem with guaranteed coverage,
    \item we present and release\footnote{\url{https://github.com/florisdenhengst/hierarchical-conformal-classification}} three hierarchical classification data sets\footnote{\url{https://huggingface.co/datasets/florisdh/hcc}}. We demonstrate the effectiveness of HCC through theoretical analysis, empirical quantitative evaluation, and a preliminary user study.
\end{enumerate}
\noindent HCC thereby bridges structured prediction and uncertainty quantification, with broad applicability across domains, offering a foundation for more reliable and user-aligned classification systems.


\section{Background}\label{sec:preliminaries} 
\paragraph{Notation.}  
We use calligraphic letters (e.g. \( \mathcal{X} \)) to denote sets, and bold lowercase letters (e.g., \( \mathbf{x} \)) for deterministic vectors. Bold uppercase letters (e.g., \( \mathbf{X} \)) denote matrices. We focus on multi-class classification problems, where the input space is \( \mathcal{X} \subset \mathbb{R}^d \), and the label space is \( \mathcal{Y} = \{1, \ldots, k\} \). A trained predictive model is denoted by \( \hat{f} : \mathcal{X} \rightarrow [0, 1]^k \) and \( \sum_{i=1}^k \hat{f}(\mathbf{x})_i = 1 \). That is, given an input \( \mathbf{x} \in \mathcal{X} \), the model outputs prediction scores, which can be a probability distribution, over the \( k \) classes. We denote the calibration and test data sets as \( \mathcal{D}_{c} \) and \( \mathcal{D}_{t} \), respectively. Each set contains $n_c$, respectively $n_t$, data pairs $(\mathbf{x}_i, y_i)$, where $y_i \in \mathcal{Y}$.




\subsection{Conformal prediction}\label{subsec:conformal-prediction}

CP equips any probabilistic classifier with a finite-sample \emph{coverage guarantee} that a prediction will contain the correct label with probability at least \( 1 - \alpha \). Given a test input \( \mathbf{x}_t \in \mathcal{D}_{t} \) and a desired maximum error rate \( \alpha \in [0,1) \), CP returns a prediction set \( \mathcal{Y}' \subseteq \mathcal{Y} \) such that \(
\mathbb{P}(y_t \in \mathcal{Y}') \geq 1 - \alpha,\) where \( y_t \in \mathcal{Y} \) is the true label of \( \mathbf{x}_t \). 

We adopt the standard \emph{split conformal prediction} approach in which the goal is to find some conformal predictor \(h_{\mathcal{Y}}\in\mathcal{X}\times [0,1) \to 2^\mathcal{Y}\)~\citep{vovk2005algorithmic,lei2018distribution}. A fixed model \( \hat{f} \) is calibrated using a separate calibration set \( \mathcal{D}_{c}\), and a \emph{non-conformity score}
\(
s \in \mathcal{X} \times \mathcal{Y} \to \mathbb{R}.
\)
An example of this score is the predicted value of class \( y \) for input \( \mathbf{x} \), i.e., 
\(
s(\mathbf{x}, y) := 1-\hat{f}(\mathbf{x})_y.
\)
For each instance in the calibration set, CP computes the nonconformity score \( s(\mathbf{x}_i, y_i) \), resulting in the set \( \mathcal{S} = \{s(\mathbf{x}_i, y_i)\}_{i=1}^{n_{c}} \). 
Next, the quantile $\hat{q}$ of the empirical distribution of $\mathcal{S}$ is determined so that the desired coverage ratio $1-\alpha$ is achieved by choosing $\hat{q}=\lceil(n_c+1)(1-\alpha)\rceil/n_c$ where $\lceil\cdot\rceil$ denotes the ceiling function\footnote{This is essentially the $1-\alpha$ quantile with a minor adjustment}. At inference time, the prediction set is constructed as
\(
C(\mathbf{x}_t) := \left\{ y \in \mathcal{Y} : s(\mathbf{x}_t, y) \leq \hat{q} \right\}\).

\subsection{Hierarchies}
We assume that the classes are organized in a hierarchy, modeled as a DAG \( \mathcal{T} := (\mathcal{V}, \prec) \) of depth \( \delta \). The node set \( \mathcal{V} \) contains all hierarchical class labels, and the base classifier classes correspond to the set of leaves \( \mathcal{Y} \subseteq \mathcal{V} \). The edge relation \( \prec \) denotes the strict “parent-of” relationship, satisfying asymmetry, anti-reflexivity, and transitivity.

\begin{definition}[Hierarchy]\label{def:hierarchical_classification}
Let \( \mathcal{T} = (\mathcal{V}, \prec) \) be a DAG where \( \prec \) is a strict partial order. We refer to \( \mathcal{T} \) as a class hierarchy over nodes \( \mathcal{V} \).
\end{definition}

\begin{wrapfigure}{r}{.5\textwidth}
\vspace{-1em}
\centering
\includegraphics[width=.5\columnwidth,trim={0.2cm 3.8cm 5.4cm .2cm}, clip]{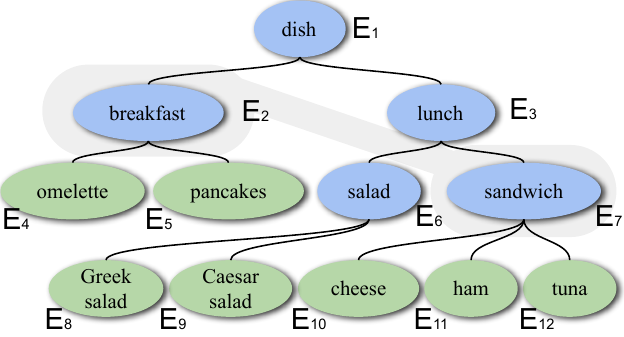}
\caption{Taxonomy for a running example.}
\label{fig:running exmp}
\vspace{-1em}
\end{wrapfigure}
We denote the transitive closure of \( \prec \), i.e. the relation indicating all ancestors of a node, by \( \precprec \), and
 assume a single  \( \mathtt{root} \in \mathcal{V}: \nexists v \in \mathcal{V}, v \prec \mathtt{root} \).

\begin{definition}[Leaves] \label{def:leaves}
Let \( \mathcal{T} = (\mathcal{V}, \prec) \) be a hierarchy. The set of \emph{leaves}, with the assumption that \( \mathcal{Y} = \text{leaves}(\mathcal{T}) \), is defined as:
\(
\text{leaves}(\mathcal{T}) := \left\{ v \in \mathcal{V} \mid \nexists w \in \mathcal{V} \text{ such that } v \prec w \right\}.
\)
\end{definition}

\begin{definition}[Lowest Common Ancestor, LCA] \label{def:lca_set}
Let \( \mathcal{T} \) be a hierarchy. For a non-empty set of nodes \( \mathcal{S} \subseteq \mathcal{V} \), define the set of \emph{common ancestors} as:
\(
\text{Anc}_{\mathcal{T}}(\mathcal{S}) := \left\{a \in \mathcal{V} \mid \forall v \in \mathcal{S},\, a \precprec v \right\}.
\)
The \emph{lowest common ancestor(s)} of \( \mathcal{S} \), \( \text{LCA}_{\mathcal{T}}(\mathcal{S}) \), is the set of maximal elements in \( \text{Anc}_{\mathcal{T}}(\mathcal{S}) \) with respect to \( \precprec \):
{\small 
\(
\text{LCA}_{\mathcal{T}}(\mathcal{S}) := \left\{ a^\star \in \text{Anc}_{\mathcal{T}}(\mathcal{S}) \mid \nexists a' \in \text{Anc}_{\mathcal{T}}(\mathcal{S}) : a^\star \precprec a' \right\}.
\)
}
\end{definition}
As a running example, we consider a classification problem over 7 dish types
in Figure~\ref{fig:running exmp}. It consists of twelve nodes in a DAG of depth four, with seven leaves highlighted in green. These leaves correspond to the classification labels. \texttt{lunch} is the lowest common ancestor of \{\texttt{Caesar salad}, \texttt{cheese sandwich}\}.

\subsection{Conformal Prediction and Hierarchies}
A learn-then-test approach has been used by the Conformal Structured Prediction (CSP) approach~\citep{zhang2024conformal}. This approach considers increasingly desirable predefined candidate thresholds $\boldsymbol{\tau} = (\tau_1, \dots, \tau_k)$ sequentially, and generates a solution for each threshold until a threshold $\tau_j$ is encountered for which the coverage is no longer guaranteed according to a specific statistical test, and returns threshold $\tau_{j-1}$ as its solution.

Although CSP is flexible enough to be applicable to both regression and classification, it comes with several theoretical and practical limitations.
First, the set of candidate thresholds \(\boldsymbol{\tau}\) has to be set upfront to contain sufficiently conservative thresholds to meet the coverage guarantee, and crucially, has to contain the true optimal threshold in order to find an optimal solution. Second, the generation of a solution for each $\tau_i \in \boldsymbol{\tau}$ requires solving an integer programming problem which scales with the size of the taxonomy, whereas our solution focuses on leveraging DAG structure to prune the search substantially. In addition, the CSP algorithm considers the candidate thresholds in order, which creates computational efficiency issues which cannot be addressed easily with parallel execution.

Another approach to applying CP in the context of hierarchical label spaces in a DAG was proposed in the context of the conformal risk control (CRC) framework~\citep{angelopoulosconformal,10.1093/jrsssc/qlag037}. This framework requires a monotonic loss $L$ that can be bound on any sample $x_{n_{c}+1}$ not seen during calibration by a single hyperparameter $\alpha$:
\(\mathbb{E}\left[ L(C(\mathbf{x}_{n_c+1}), y_{n_c+1}) \right] \leq \alpha.\)
As far as we are aware, no monotone loss can be formulated to explicitly control the trade-off between nominal prediction set size and leaf coverage as we intend to do in this work. Alternatively,~\citet{angelopoulosconformal} bound the  graph distance between the selected interior node and the closest ancestor of the ground truth 
\(
L_{t}(C(\mathbf{x}_t), y) = d(C(\mathbf{x}_t), y_t) \),
where $d(C(\mathbf{x}_t), y_t)$ denotes the smallest number of edges between the nodes in the prediction set and the ground truth. This approach predicts a single node whereas our approach flexibly generates prediction sets of varying sizes and containing both internal and leaf nodes adaptively based on the underlying classifiers' predicted scores, as well as the graph structure.

\section{Hierarchical Conformal Classification} 
\label{sec:methodology}


We consider the problem of producing compact and informative prediction sets with coverage guarantees,  under the assumption that class labels are structured according to a known DAG. Each prediction set should ideally contain a small number of nodes to promote usability and avoid redundancy. Moreover, each leaf represents a concrete class label, and the prediction of an internal node with many leaves as its descendants implies greater uncertainty of the base classifier. Therefore, the nodes in the prediction set should have a small number of leaves as descendants. Thus, compact prediction sets that have a small number of leaves as descendants provide more specific and actionable predictions. These objectives must be achieved while still satisfying the standard conformal prediction coverage guarantee to ensure the reliability of the prediction.

To formalize these desiderata, we introduce the notion of a leaf cover, which quantifies the implied coverage of class nodes for a set of arbitrary nodes $\mathcal{N}$.
\begin{definition}[Leaf Cover] \label{def:leaf-cover}
For any subset \( \mathcal{N} \subseteq \mathcal{V} \) of a DAG \(\mathcal{T}=(\mathcal{V},\prec)\), the \emph{leaf cover} of \( \mathcal{N} \), denoted \( \text{leaf-cov}_\mathcal{T}(\mathcal{N}) \), is defined as:
\[
\bigcup_{v \in \mathcal{N}} \left\{ w \in \text{leaves}(\mathcal{T}) \mid v \precprec w \text{ or } w = v \right\},
\]
\end{definition}
i.e. the union of leaves reachable from all nodes in \(\mathcal{N}\). In Figure~\ref{fig:running exmp}, \{\texttt{breakfast}, \texttt{sandwich}\} has leaf cover \{\texttt{omelette}, \texttt{pancakes}, \texttt{cheese sandwich}, \texttt{ham sandwich}, \texttt{tuna sandwich}\}.

Based on the definition of a leaf cover, we put forward our goal to construct a hierarchical conformal predictor \( h^*_{\beta, \mathcal{V}} : \mathcal{X} \times [0,1) \to 2^{\mathcal{V}} \) that returns a set of nodes for any test pair \((\mathbf{x}_t, y_t) \sim \mathcal{D}_t\), such that:
\begin{align}
&h^*_{\beta, \mathcal{V}}(\mathbf{x}_t, \alpha) = \underset{\mathcal{N} \subseteq \mathcal{V}}{\arg\min}~|\mathcal{N}| + \beta \left| \bigcup_{v \in \mathcal{N}} \mathrm{leaf\text{-}cov}_{\mathcal{T}}(\{v\}) \right| \nonumber \\
&\;\text{subject to}~ \mathbb{P}\Big(y_t \in \mathrm{leaf\text{-}cov}_{\mathcal{T}}(\mathcal{N})\Big) \geq 1 - \alpha.
\label{eq:objective}
\end{align}
Here, \( \beta \in \mathbb{R}_+ \) controls the trade-off between compactness and specificity, quantified as the number of covered leaves. A higher \( \beta \) prefers solutions that covers fewer leaves. Such solutions tend to include more nodes located closer to the leaves.
The minimization objective in equation~\eqref{eq:objective} expresses that the set \( \mathcal{N} \) must be compact and semantically meaningful, and the constraint guarantees that its leaf cover includes the ground-truth label with probability $1-\alpha$.



At the heart of HCC is this constrained optimization problem, which formalizes the goal of producing valid and semantically coherent prediction sets under a given label hierarchy. However, solving this problem exactly is infeasible, as the number of candidate solutions grows combinatorially with the number of nodes in the DAG. To address this, HCC combines learning with efficient search strategies that exploit the hierarchical structure of the DAG. Specifically, we first calibrate several conformal classifiers over a hierarchical label space, and then use these calibrated classifiers to guide an efficient inference-time procedure that approximates the optimal solution. Marginal coverage guarantees are maintained in the face of simultaneous inference and test-time selection using an efficient Westfall-Young MaxT procedure~\citep{westfall1993resampling}.


\subsection{Calibrating Hierarchical Classifiers}
We consider the calibration of HCC in three phases, illustrated in Figure \ref{fig:hcp_training}. Phase one  in Figure~\ref{fig:hcp_training}(b) consists of constructing a reduced search space. Instead of enumerating all node subsets, we restrict our attention to a much smaller family of subsets without affecting the optimality of the solution. Phase two in Figure~\ref{fig:hcp_training}(c) comprises the propagation of scores and labels. For every candidate \(\mathcal{S}\), we propagate the labels and base classifier's predicted scores from the leaf level upward, assigning to each internal node the logical OR of its leaf cover labels and the sum of the base classifier's predictions. Finally, in phase three, we calibrate a conformal predictor on each subset with the propagated labels and the propagated base classifier scores resulting in a collection of calibrated predictors as depicted in Figure~\ref{fig:hcp_training}(d). This collection constitutes our model family:
$
\mathcal{H} \;=\; \bigl\{\,h_{\beta, \mathcal{S}}:\mathcal{S}\subseteq \mathcal{V}\bigr\}.
$
We now elaborate on these three steps.

\paragraph{Search-Space Pruning}
\label{sec:NOL-covers}
The objective in equation~\eqref{eq:objective} involves a minimization over all subsets $\mathcal{S} \subseteq \mathcal{V}$, which is of size \(2^{|\mathcal{V}|}\) and computationally infeasible for all but the smallest hierarchies. To address this, we propose an effective pruning of the search space based on two structural insights about the hierarchy. Specifically, we restrict our search to subsets of nodes that (i) collectively cover all leaves in the hierarchy $\mathcal{Y}$, and (ii) do not include any ancestor–descendant pairs. We present these two notions in the following definition of \emph{non-overlapping leaf covers}, and then show why these two desiderata are useful for efficient search.

\begin{definition}[NOL-cover]
\label{def:nol-cover}
A subset \( \mathcal{S} \subseteq \mathcal{V} \) is called a \emph{non-overlapping leaf cover} (NOL-cover) if:
\begin{enumerate}
    \item \textbf{Leaf coverage}  all leaves are covered by at least one node in \( \mathcal{S} \) via their descendants:
    \(
    \text{leaf-cov}_\mathcal{T}(\mathcal{S}) = \mathcal{Y}
    \), and
    \item \textbf{Ancestral independence} no two nodes in \( \mathcal{S} \) are related by an ancestor-descendant relationship:
    \(
    \forall u, v \in \mathcal{S},\ u \neq v \Rightarrow \neg(u \precprec v) \land \neg(v \precprec u).
    \)
\end{enumerate}
\end{definition}

For the running example in Figure~\ref{fig:running exmp}, the set \{\texttt{breakfast}, \texttt{sandwich}\} fails to satisfy the leaf coverage requirement and therefore is not a valid NOL-cover. It could be extended to an NOL-cover by extending it with either \{\texttt{salad}\} or \{\texttt{Greek salad}, \texttt{Caesar salad}\}, but not both simultaneously, as this would violate the ancestral independence requirement.

We show that every conformal predictor optimal to problem~\eqref{eq:objective} must be trained on a set that satisfies the leaf coverage requirement for any non-trivial $\alpha$, since it would violate the probabilistic coverage requirement otherwise. We additionally show that the ancestral independence requirement excludes solutions that are suboptimal to~\eqref{eq:objective}. All proofs are in Appendix~\ref{app:proofs}.

\begin{restatable}[Necessity of leaf-coverage]{proposition}{PropCoverage}
\label{prop:necessity}
Any  solution $h_{\cdot,\mathcal{S}}$ feasible to equation~\eqref{eq:objective} must cover all leaves in \( \mathcal{Y} \) for any $\alpha \leq \underset{y\in \mathcal{Y}}{\min}~\mathbb{P}(Y=y)$; that is,
$
\text{leaf-cov}_{\mathcal{T}}(\mathcal{S}) = \mathcal{Y}
$ for any non-trivial $\alpha$.
\end{restatable}
\begin{restatable}[Necessity of ancestral independence]{proposition}{PropAncestral}
\label{prop:sufficiency}
Any solution to problem~\eqref{eq:objective} is necessarily ancestral independent.
\end{restatable}

These propositions suggest that it suffices to restrict the search space to NOL-covers only, thereby substantially reducing the search space in practice. The particular size of the reduction depends on the DAG, e.g. the presence of multiple ancestry and its branching factor. However, in a worst case balanced binary tree of depth $d$, the search space remains in approximately $\mathcal{O}(1.5028^{2^d})$ (see Appendix~\ref{app:search_space_analysis}). In cases where the full space of NOL covers is intractable, a limited search can be employed at the cost of optimality per level of the tree, where a candidate NOL cover is constructed at each depth $d_i \leq d$ with all nodes at that depth $d_i$ and all leaves at a lower depth $d_j \leq d_i$. In this case, optimality is not guaranteed but coverage guarantees are maintained because each thus constructed candidate is a NOL cover. In other cases, we structurally prune suboptimal solutions from $\mathcal{H}$ by only considering the NOL-covers $\mathcal{H} := \{h_{\beta, \mathcal{S}} | \mathcal{S} \in \text{NOL-covers} \}$ and further reduce the search during inference as detailed in Section~\ref{sec:inference}.

\paragraph{Propagating Scores and Labels}
\label{sec:propagate}
Each conformal classifier in $\mathcal{H}$ is calibrated with respect to a specific subset of nodes in the DAG, including internal nodes. However, the classification scores $\hat{f}$ and the ground-truth labels are available only for leaves. To enable calibration at internal nodes, the classifier scores must be propagated upward. To do so, we sum the scores of all descendant leaves for each node and mark all ancestors of the true leaf as the ground truth label. This propagation ensures that every node in the DAG is associated with a predicted classification score and ground-truth label for each instance.
\begin{definition}[Propagated score]
For any node \( v \in \mathcal{V} \) the \emph{propagated score} of \( v \) is defined as:
\[
\hat{g}(\mathbf{x})_v := \sum_{w \in \text{leaf-cov}_{\mathcal{T}}(\{v\})} \hat{f}(\mathbf{x})_w,
\]
i.e., the sum of the predicted scores of all leaf descendants of \( v \), according to the base classifier \( \hat{f} \).
\end{definition}
\begin{definition}[Propagated label]
For any input-label pair \( (\mathbf{x}_i, y_i) \), the propagated ground-truth set for that instance $i$ is defined as
\(
\mathcal{Y}'_i := \left\{ v \in \mathcal{V} \mid y_i \in \text{leaf-cov}_{\mathcal{T}}(\{v\}) \right\},
\)
i.e., all ancestors of an instance's ground-truth leaf node form its hierarchical ground-truth labels.
\end{definition}





If the hierarchy \(\mathcal{T}\) is a tree, then there is at most one ground-truth node per NOL-cover for each test instance. However, in the more general case where \( \mathcal{T} \) is a DAG but not a tree, multiple nodes in a NOL-cover may be in an instance's ground-truth label set. Consequently, the CP task for a NOL-cover requires procedures tailored to the multi-label setting.
In multi-label classification, the ground truth labels can be represented by a vector indicating ground truths. We here define such a ground-truth vector per NOL-cover, i.e. for a NOL-cover of size $l$, we define an associated indicator vector \(\mathbf{y}_i\in\{0,1\}^l\) where \(\mathbf{y}_{i,j}=1\) if class \(j \in \mathcal{Y}'_i \) is relevant for example $i$ and 0 otherwise. We provide details of the calibration and inference for this setting below.


\paragraph{Calibration}
\label{sec:calibration}
During calibration, a conformal predictor is constructed for each NOL-cover in the search space. Each predictor is then calibrated using propagated scores and labels as detailed previously. The calibration procedure follows the general split-CP process of estimating a quantile $\hat{q}$ on a calibration set of nonconformity scores. In particular, in cases where $\mathcal{T}$ is a tree, any preferred CP approach can be used. However, for the more general case where $\mathcal{T}$ is a DAG, the calibration problem is more challenging due to the multi-label nature of the problem. We propose a novel nonconformity score function to address this challenge.

The proposed nonconformity score function is based on the insight that multiple true labels correspond to a single ground-truth leaf and that it suffices to include only one such label to satisfy the coverage condition in equation~\eqref{eq:objective}. Intuitively, a nonconformity score should be low when the classifier puts most of its probability mass on a node of
$\mathcal{S}$ that contains the true class. Since any such node suffices for coverage, only the node the classifier is most confident about determines the score.
 We formalize this insight in a score inspired by~\citet{sadinle2019least}:
\begin{equation}
\label{eq:hcp_score}
s(\mathbf{x}_i, \mathbf{y}_i, \mathcal{S}) = 1-\hat{g}(\mathbf{x}_i)_{v'_i}
\end{equation}
where \(v'_i \in \mathcal{S}\) is the label with the highest classifier score $
v'_i := \underset{{j : \mathbf{y}_{i,j} = 1}}{\arg\max}~\hat{g}(\mathbf{x}_i)_j$. 

We note that this score trivially reduces to the selection of the prediction score for the ground truth label in the single-label setting. Therefore, $s$ can be seen as a generalization of the score proposed by~\citet{sadinle2019least} to a setting with hierarchically derived labels and scores.

We now show that the nonconformity score in equation~\eqref{eq:hcp_score} produces prediction sets with hierarchical coverage guarantees. In particular, it ensures that for any test instance \(\mathbf{x}_t\) and any allowed error rate \(\alpha \in [0,1)\), the probability that its ground truth leaf label \(y_t\) is a descendant of one of the nodes included in the prediction set is \(\geq 1-\alpha\). All proofs are in Appendix~\ref{app:proofs}.
\begin{restatable}[Hierarchical Coverage Guarantee]{theorem}{CoverageTheorem}
\label{thm:coverage}
Let $\mathcal{S}\subseteq \mathcal{V}$ be a NOL-cover, \((\mathbf{x}_{t}, y_{t})\sim \mathcal{D}_t\) be an exchangeable test instance, \(\mathbf{y}_t\) the hierarchical ground truth label indicator for $\mathcal{S}$ and $y_t$.
Construct a prediction set as follows using the nonconformity score function $s$ from equation~\eqref{eq:hcp_score}:
\[
C(\mathbf{x}_{t}, \mathcal{S}) := \left\{ v \in \mathcal{S} \mid s(\mathbf{x}_{t}, \cdot, \mathcal{S}) \leq \hat{q} \right\},
\]
where \(\hat{q}\) is the empirical quantile \(\lceil(n_c+1)(1-\alpha)\rceil/n_c\) of the calibration nonconformity scores
\(
\{s(\mathbf{x}_i, \mathbf{y}_i, \mathcal{S})\}_{i=1}^{n_{c}}
\) obtained during calibration.
Then, the prediction set satisfies the hierarchical coverage guarantee
\[
\mathbb{P} \left[y_{t} \in \text{leaf-cov}_T\bigl(C(\mathbf{x}_{t}, \mathcal{S})\bigr)\right] \geq 1 - \alpha, \, \forall \alpha \in [0,1).
\]
\end{restatable}
\noindent The nonconformity score function can be further adapted to handle scenarios such as adaptive calibration, large or grouped label sets, and the related coverage guarantee proofs can be adapted analogously~\citep{angelopoulos2020uncertainty,ding2023class}.

\begin{table*}[t]
\centering\caption{Test set results (mean$\pm$SD). \textbf{Bold} indicates meeting coverage guarantee for `coverage' column, and, for other columns, the best out of all methods meeting the coverage guarantee.
}
\label{tab:base_results}
\setlength{\tabcolsep}{3pt}
\small
\begin{tabular}{l l l l r@{\hskip 2pt}l r@{\hskip 2pt}l r@{\hskip 2pt}l r@{\hskip 2pt}l}\toprule
      & $\beta$ & $1-\alpha$ & method           & \multicolumn{2}{c}{coverage $\uparrow$} 
                     & \multicolumn{2}{c}{costs $\downarrow$} 
                     & \multicolumn{2}{c}{PS size $\downarrow$} 
                     & \multicolumn{2}{c}{covered leaves $\downarrow$} \\
   \midrule
 \texttt{dbp} & .19   & .9997 & Standard CP                    & \textbf{.9998} & $\pm$ .02 &      3.18 & $\pm$ 2.62  &       2.67 & $\pm$ 2.20  &      \textbf{2.67} & $\pm$ 2.20     \\
   & &  & LCA                        & \textbf{.9998} & $\pm$ .02 &      2.20  & $\pm$ 1.13   &       \textbf{1.00}    & $\pm$ .00  &      6.32 & $\pm$ 5.94     \\
   & &  & CSP                        & .9938 & $\pm$ .08 &      1.20  & $\pm$ .15  &       1.01 & $\pm$ .09  &      1.02 & $\pm$ .50     \\[.3ex]
   & &  & HCC (ours)              & \textbf{.9999} & $\pm$ .01 &      \textbf{2.01} & $\pm$ .94  &       1.23 & $\pm$ .45  &      4.12 & $\pm$ 4.47     \\
   & &  & ~~w/o dynamic pruning      & \textbf{.9999} & $\pm$ .01 &      2.15 & $\pm$ 1.03  &       1.34 & $\pm$ .62  &      4.30 & $\pm$ 4.61     \\
   & &  & ~~w/o correction           & .9997 & $\pm$ .02 &      1.88 & $\pm$ .85  &       1.22 & $\pm$ .43  &      3.44 & $\pm$ 3.69     \\
   & &  & ~~w/o score                & \textbf{.9999} & $\pm$ .01 &      2.09 & $\pm$ 1.02  &       1.28 & $\pm$ .59  &      4.28 & $\pm$ 4.59     \\[1ex]
 \texttt{img} & .04 & .98  & Standard CP                     & \textbf{.9807} & $\pm$ .14 &     11.49 & $\pm$ 20.49 &      11.04 & $\pm$ 19.70 &     \textbf{11.04} & $\pm$ 19.70    \\
  & &   & LCA                        & \textbf{.9807} & $\pm$ .14 &     14.79 & $\pm$ 15.63  &       \textbf{1.00}    & $\pm$ .00  &    344.69 & $\pm$ 390.64   \\
  & &   & CSP                        & \textbf{.9850}  & $\pm$ .12 &     15.98 & $\pm$ 17.03 &       5.03 & $\pm$ 2.98  &    269.83 & $\pm$ 364.12   \\[.3ex]
  & &   & HCC (ours)              & \textbf{.9882} & $\pm$ .11 &      \textbf{7.17} & $\pm$ 10.26 &       3.73 & $\pm$ 5.18  &     85.85 & $\pm$ 221.61   \\
  & &   & ~~w/o dynamic pruning      & \textbf{.9881} & $\pm$ .11 &      7.33  & $\pm$ 10.51 &       3.99 & $\pm$ 5.43  &     83.55 & $\pm$ 221.77   \\
  & &   & ~~w/o correction           & .9322  & $\pm$ .25 &      3.11 & $\pm$ 3.93  &       1.70 & $\pm$ 1.57  &     35.22 & $\pm$ 78.06   \\
  & &   & ~~w/o score                & .9531 & $\pm$ .21 &      2.67  & $\pm$ 3.09  &       2.29 & $\pm$ 2.73  &      9.37 & $\pm$ 22.12    \\[1ex]
 \texttt{gtz} & .25 & .9 & Standard CP & \textbf{0.9175} & $\pm$ 0.28 &      4.07 & $\pm$ 2.34  &       3.25 & $\pm$ 1.87 &      \textbf{3.25} & $\pm$ 1.87     \\
  & &   & LCA                        & \textbf{0.9175} & $\pm$ 0.28 &      2.87 & $\pm$ 1.00     &       \textbf{1.00}    & $\pm$ .00 &      7.47 & $\pm$ 4.01     \\
  & &   & CSP                        & \textbf{.9762} & $\pm$ .15 &      4.67 & $\pm$ 1.67  &       3.30  & $\pm$ 1.09  &      5.53 & $\pm$ 2.73     \\[.3ex]
  & &   & HCC (ours)              & \textbf{.9275} & $\pm$ .26 &      \textbf{2.47} & $\pm$ .94  &       1.32 & $\pm$ .47  &      4.61 & $\pm$ 3.80     \\
  & &   & ~~w/o dynamic pruning      & \textbf{1.}      & $\pm$ .00 &      3.41 & $\pm$ .35  &       1.07 & $\pm$ .26  &      9.35 & $\pm$ 2.11     \\
  & &   & ~~w/o correction           & \textbf{.9275} & $\pm$ .26 &      \textbf{2.47} & $\pm$ .94  &       1.33 & $\pm$ .47  &      4.56 & $\pm$ 3.75     \\
  & &   & ~~w/o score                & \textbf{.9575} & $\pm$ .20 &      2.67 & $\pm$ .97  &       1.18 & $\pm$ .38  &      5.97 & $\pm$ 4.22     \\
\bottomrule
\end{tabular}
\end{table*}



\subsection{Inference with Dynamic Pruning}
\label{sec:inference}
We now detail how the family of predictors is used to produce a prediction for a test instance \(\mathbf{x}_t\) and a target error level \(\alpha \in [0,1)\) as visualized in Figure~\ref{fig:hcp_inference}. We follow a three-step procedure. Classifier output $\hat{f}$ (\ref{fig:hcp_inference}a) is propagated upward to compute the scores \(\hat{g}(\mathbf{x}_{\mathrm{t}})\) for all nodes, analogously to the propagation of the scores in the calibration phase (\ref{fig:hcp_training}c). Each conformal classifier \(h_i \in \mathcal{H}\) is then evaluated on these scores with an adjusted \(\alpha\), and produces a candidate prediction set (\ref{fig:hcp_inference}c). The candidate that minimizes the cost in~\eqref{eq:objective} is selected (\ref{fig:hcp_inference}d) to effectively form \(h^*_{\beta,\mathcal{V}}\) for that instance $\mathbf{x}_t$.

Since inference involves evaluating multiple conformal classifiers in parallel, we need to correct the user-selected $\alpha$ to account for multiple comparisons. We therefore derive a corrected $\alpha'$ using an approach analogous to the single-step Westfall-Young MaxT procedure~\citep{westfall1993resampling}. This procedure provides exact and tight family-wise error rate control using the dependence structure between the underlying test statistics, without requiring the a priori sorting needed for multi-step methods, based on the maximum. In our approach, for each calibration instance $i$ we compute the nonconformity scores of all $m$ predictors,
$\mathbf{S}_i := (s_{i,1},\dots,s_{i,m})$, and set $t_i := \max(\mathbf{S}_i)$. Every predictor then uses
the same threshold $\hat{q}$, the $\lceil(n_c+1)(1-\alpha)\rceil/n_c$ empirical quantile of
$\{t_i\}_{i=1}^{n_c}$. By exchangeability, $\mathbb{P}(t_t \leq \hat{q}) \geq 1-\alpha$, so with
probability at least $1-\alpha$ \emph{every} predictor in $\mathcal{H}$ covers $y_t$ simultaneously,
and hence so does any prediction set selected from them.

We further prune the search based on the insight that any optimal solution to~\eqref{eq:objective} should be an intermediate of the predictor on the leaves, and an element of the LCA-set of its resulting prediction set. 
\begin{restatable}[Pruning intermediate solutions]{proposition}{PropPrune}
\label{prop:prune}
For any test instance $\mathbf{x}_t$, and associated predictor $h^*_{\beta,\mathcal{S}}$ optimal according to~\eqref{eq:objective} defined over a NOL-cover $\mathcal{S}$ such that $\exists v \in \mathcal{S}, v \precprec lca, lca \in LCA\left(C\left(\mathbf{x}_t, \mathcal{Y}\right)\right)$, any $lca \in LCA(C(\mathbf{x}_t,\mathcal{Y}))$ over the leaves is also optimal.
\end{restatable}
We therefore prune all NOL-covers that contain an ancestor of $LCA\left(C\left(\mathbf{x}_t,\mathcal{Y}\right)\right)$, further reducing the search during inference without risking to eliminate the optimal solution given $\mathcal{H}$.
This pruning also suggests an efficient form of the Westfall-Young MaxT correction. To see this, first note that a NOL cover consisting of only leaves $\mathcal{Y}$ is always in $\mathcal{H}$. Any node's propagated prediction is at least that of its leaf descendants, so its nonconformity scores $s_{i,\cdot}$ are always less than or equal to the associated score for the leaf-only cover. The corrected threshold thus equals the standard CP threshold on the leaves, and the leaf-level prediction set is
unaffected by the correction.


\newlength{\axeswidth}
\setlength{\axeswidth}{0.32\textwidth}
\begin{figure*}[t]
    \centering
    \begin{subfigure}[t]{.32\textwidth}
        \centering
        \begin{tikzpicture}

\begin{axis}[
    width=\axeswidth,
    axis y line*=left,
    axis x line*=none,
    xlabel={$\beta$},
    ylabel={PS size},
    ymin=1, ymax=2.0,
    xmin=0, xmax=1,
    legend pos=north west,
    grid=both,
    xlabel near ticks,
    ylabel near ticks,
    enlargelimits=false,
    every axis x label/.style={at={(current axis.south)},anchor=north, below=3mm},
    x axis line style={black},
    y axis line style={cbvermillion},
    ylabel style={cbvermillion, at={(ticklabel cs:0.5)}, yshift=-5pt, anchor=south},
    yticklabel style={cbvermillion},
]
\draw[dotted, thick, red] (axis cs:0.19,1) -- (axis cs:0.19,5.1);
\addplot+[cbvermillion, thick, mark=*, mark options={fill=cbvermillion}] table[row sep=\\] {
beta value\\
0.0    1.000000     \\
0.1    1.163036     \\
0.2    1.310964     \\
0.3    1.354357     \\
0.4    1.451696     \\
0.5    1.513554     \\
0.6    1.542286     \\
0.7    1.558071     \\
0.8    1.574643     \\
0.9    1.577750     \\
1.0    1.851946         \\
};
\end{axis}

\begin{axis}[
    axis y line*=right,
    axis x line=none,
    ylabel={\# covered leaves},
    ymin=1, ymax=6.4,
    xmin=0, xmax=1,
    legend pos=north east,
    color=cbblue,
    xlabel near ticks,
    ylabel near ticks,
    enlargelimits=false,
    ylabel style={cbblue, at={(ticklabel cs:0.5)}, yshift=-10pt, anchor=south},
    width=\axeswidth,
]
\addplot+[cbblue, thick, mark=square*, mark options={fill=cbblue}] table[row sep=\\] {
beta    value\\
0.0    6.319500     \\
0.1    4.542679     \\
0.2    3.693946     \\
0.3    3.523464     \\
0.4    3.234286     \\
0.5    3.105964     \\
0.6    3.057536     \\
0.7    3.033500     \\
0.8    3.010875     \\
0.9    3.007161     \\
1.0    2.732964     \\
};
\end{axis}
\end{tikzpicture}
        \vspace{-1.6em}
        \caption{\texttt{dbp}}
        \label{fig:betas_dbpedia}
    \end{subfigure}\hfill
    \begin{subfigure}[t]{.32\textwidth}
        \centering
        \begin{tikzpicture}

\begin{axis}[
    width=\axeswidth,
    axis y line*=left,
    axis x line*=none,
    xlabel={$\beta$},
    ylabel={PS size},
    ymin=1, ymax=11,
    xmin=0, xmax=1,
    legend pos=north west,
    grid=both,
    xlabel near ticks,
    ylabel near ticks,
    enlargelimits=false,
    every axis x label/.style={at={(current axis.south)},anchor=north, below=3mm},
    x axis line style={black},
    y axis line style={cbvermillion},
    ylabel style={cbvermillion, at={(ticklabel cs:0.5)}, yshift=-5pt, anchor=south},
    yticklabel style={cbvermillion}
]
\addplot+[cbvermillion, thick, mark=*, mark options={fill=cbvermillion}] table[row sep=\\,    width=0.33\linewidth] {
    beta    value \\
0.00      1.000000       \\
0.01      1.656575       \\
0.02      2.368400       \\
0.03      3.057625       \\
0.04      3.735075       \\
0.05      4.419575       \\
0.06      5.035025       \\
0.07      5.629500       \\
0.08      6.166250       \\
0.09      6.639175       \\
0.10      7.058825       \\
0.20      9.485675       \\
0.30     10.066475       \\
0.40     10.320925       \\
0.50     10.499250       \\
0.60     10.555325       \\
0.70     10.606500       \\
0.80     10.659025       \\
0.90     10.677075       \\
1.00     10.781075       \\
};
\draw[dotted, thick, red] (axis cs:0.04,1) -- (axis cs:0.04,5.1);
\end{axis}

\begin{axis}[
    axis y line*=right,
    axis x line=none,
    ylabel={\# covered leaves},
    ymin=1, ymax=350,
    xmin=0, xmax=1,
    legend pos=north east,
    color=cbblue,
    xlabel near ticks,
    ylabel near ticks,
    enlargelimits=false,
    ylabel style={cbblue, at={(ticklabel cs:0.5)}, yshift=-10pt, anchor=south},
    width=\axeswidth,
]
\addplot+[cbblue, thick, mark=square*, mark options={fill=cbblue}] table[row sep=\\,    width=0.33\linewidth] {
    beta    value \\
0.00    344.695175       \\
0.01    182.187875       \\
0.02    133.195775       \\
0.03    105.308025       \\
0.04     85.809625       \\
0.05     70.527225       \\
0.06     59.303800       \\
0.07     50.157600       \\
0.08     42.997950       \\
0.09     37.425600       \\
0.10     32.991575       \\
0.20     15.033500       \\
0.30     12.619125       \\
0.40     11.889025       \\
0.50     11.506925       \\
0.60     11.407825       \\
0.70     11.330100       \\
0.80     11.260175       \\
0.90     11.239100       \\
1.00     11.134650       \\
};
\end{axis}
\end{tikzpicture}
        \vspace{-1.6em}
        \caption{\texttt{img}}
        \label{fig:betas_imagenet}
    \end{subfigure}\hfill
    \begin{subfigure}[t]{.32\textwidth}
        \centering
        \begin{tikzpicture}

\begin{axis}[
    width=\axeswidth,
    axis y line*=left,
    axis x line*=none,
    xlabel={$\beta$},
    ylabel={PS size},
    ymin=1, ymax=2.5,
    xmin=0, xmax=1,
    legend pos=north west,
    grid=both,
    xlabel near ticks,
    ylabel near ticks,
    enlargelimits=false,
    every axis x label/.style={at={(current axis.south)},anchor=north, below=3mm},
    x axis line style={black},
    y axis line style={cbvermillion},
    ylabel style={cbvermillion, at={(ticklabel cs:0.5)}, yshift=-5pt, anchor=south},
    yticklabel style={cbvermillion},
]
\draw[dotted, thick, red] (axis cs:0.25,1) -- (axis cs:0.25,5.1);
\addplot+[cbvermillion, thick, mark=*, mark options={fill=cbvermillion}] table[row sep=\\] {
beta value\\
0.0    1.00000      \\
0.1    1.00000      \\
0.2    1.11250      \\
0.25   1.1125       \\
0.3    1.43500      \\
0.4    1.43500      \\
0.5    1.94500      \\
0.6    1.94500      \\
0.7    1.94500      \\
0.8    2.48500      \\
0.9    2.48500      \\
1.0    2.49750      \\
};
\end{axis}

\begin{axis}[
    axis y line*=right,
    axis x line=none,
    ylabel={\# covered leaves},
    ymin=1, ymax=8,
    xmin=0, xmax=1,
    legend pos=north east,
    color=cbblue,
    xlabel near ticks,
    ylabel near ticks,
    enlargelimits=false,
    ylabel style={cbblue, at={(ticklabel cs:0.5)}, yshift=-10pt, anchor=south},
    width=\axeswidth,
]
\addplot+[cbblue, thick, mark=square*, mark options={fill=cbblue}] table[row sep=\\] {
beta    value       \\
0.0    7.47250      \\
0.1    7.47250      \\
0.2    6.57250      \\
0.25   6.57         \\
0.3    5.44375      \\
0.4    5.44375      \\
0.5    4.42375      \\
0.6    4.42375      \\
0.7    4.42375      \\
0.8    3.74875      \\
0.9    3.74875      \\
1.0    3.73625      \\
};
\end{axis}
\end{tikzpicture}
        \vspace{-1.6em}
        \caption{\texttt{gtz}}
        \label{fig:betas_gtzan}
    \end{subfigure}
    \caption{Efficiency expressed in the {\color{cbvermillion} set sizes} and {\color{cbblue} number of covered leaves} for varying $\beta$. Dotted vertical lines indicate $\beta$ values in Table~\ref{tab:base_results}.}
    \label{fig:betas}
\end{figure*}

\section{Experiments} 
\label{sec:experiments}
\begin{wraptable}{r}{.4\textwidth}
\centering
\caption{Overview of data sets.}
\label{table:taxonomy-description}
\centering
\small
\setlength{\tabcolsep}{3pt}
\begin{tabular}{lccc}
\toprule
\textbf{}  & \texttt{dbp} & \texttt{img} & \texttt{gtz} \\ \midrule
$n$  & 70,000 & 50,000 & 1,000\\
\# labels  & 14 & 1000 & 10 \\
modality & text & image & audio \\[1ex]
classifier            & BERT & ResNet-152 & XGBoost \\
accuracy            & .99 &  .79 & .63 \\[1ex]
\# nodes  & 25 & 1860 & 15\\
\# leaves  & 14 & 1000 & 10 \\
depth  & 6 & 17 & 3\\ 

\bottomrule
\end{tabular}
\vspace{-2em}
\end{wraptable}
To evaluate our approach across modalities and classification paradigms, we propose \textit{three distinct benchmark datasets}, each paired with a structured taxonomy and a representative base classifier as summarized in Table \ref{table:taxonomy-description}. 

\textbf{DBpedia14 (\texttt{dbp})} A \emph{text classification} task built on Wikipedia abstracts, using a subset of the DBpedia taxonomy. We employed a BERT-based classifier~\citep{zhang2015character} and derived the class hierarchy from DBPedia~\citep{mendes2012dbpedia}.\vspace{.3em}
\\
\textbf{ImageNet1K (\texttt{img})} An \emph{image dataset} with 1K object categories organized under WordNet hierarchy~\citep{russakovsky2015imagenet,miller1995wordnet} with prediction scores obtained from a ResNet classifier appropriate for the visual domain, and uses the per-level fallback instead of the full DAG~\citep{he2016deep}.\vspace{.3em}
\\
\textbf{GTZAN Genre Collection (\texttt{gtz})} An \emph{audio classification} dataset containing 1,000 music clips across 10 genres, structured into a hierarchical taxonomy as introduced by~\citet{sturm2012analysis}. This dataset enables testing our method in the acoustic domain.

%
For all datasets, we used a 80/20 calibration/test split and report mean $\pm$ standard deviation calculated across all test set examples. 
Parameters of the approach were set as follows: $\alpha$ was set based on the hardness of the classification task, i.e. inversely proportional to the accuracy of the base classifier for the main results, and varied for additional analysis, see Figure~\ref{fig:additional_alpha}. For the main results, $\beta$ was established on the basis of the median number of covered leaves for every node in the DAG and was varied to assess robustness to this parameter, see Figure~\ref{fig:betas}:
\begin{equation}
    \label{eq:beta_median}
    \beta := \left(\text{median}\bigl[|\{\text{leaf-cov}_{\mathcal{T}}\left(\{v\}\right)|v \notin \text{leaves}(\mathcal{T})|\}\bigr]_{v\in\mathcal{V}}\right)^{-1}
\end{equation}

\paragraph{Baselines and Ablations}
The primary related work we compare to is Conformal Structured Prediction (CSP)~\citep{zhang2024conformal}, with hyperparameter $M$ optimized for costs. Additional baselines include \textbf{Standard CP}, which predicts on the leaves \(\mathcal{Y}\) only, and an approach that post-processes standard CP predictions, by taking a lowest common ancestor (\textbf{LCA}). This ablation predicts internal nodes only if the standard CP prediction sets contain two or more elements and heuristically approximates the objective in~\ref{eq:objective}. 
Furthermore, we ablate \textbf{dynamic pruning}, 
we ablate $\alpha$ \textbf{correction} altogether by using the original $\alpha$, and we ablate the hierarchical nonconformity \textbf{score} in equation~\eqref{eq:hcp_score} by replacing it with a multi-label recall loss in the CRC framework.

\paragraph{Results}
\begin{figure}[tbp]
\begin{tikzpicture}
\begin{axis}[
    title={Coverage vs $1-\alpha$},
    xlabel={$1-\alpha$},
    ylabel={Coverage},
    xmin=0.65, xmax=1.0,
    ymin=0.4, ymax=1.1,
    xtick={.7, .8, .85, .9, .95, .98, .99},
    xticklabels={.7, .8, .85, .9, .95, .98, .99},
    xticklabel style={
        yshift={mod(\ticknum,2) == 0 ? "0pt" : "-10pt"},
        font=\small
    },
    legend pos=south east,
    legend columns=2,
    legend style={nodes={scale=0.6, transform shape}},
    grid=major,
    width=.32\textwidth,
    height=5.5cm
]

\addplot[gray, thick, dotted, domain=0.7:1.0, forget plot] {x};

\addplot[color=cbvermillion, solid, mark=*, error bars/.cd, y dir=both, y explicit, error bar style={opacity=0.4, thin}] 
coordinates { (0.7, 0.6800) +- (0, 0.4668) (0.8, 0.8762) +- (0, 0.3295) (0.85, 0.9363) +- (0, 0.2445) (0.9, 0.9613) +- (0, 0.1931) (0.95, 0.9812) +- (0, 0.1357) (0.98, 0.9925) +- (0, 0.0863) (0.99, 1.0000) +- (0, 0.0) };
\addlegendentry{\texttt{gtz} (hcc)}

\addplot[color=cbvermillion, dashed, mark=o, error bars/.cd, y dir=both, y explicit, error bar style={opacity=0.4, thin}] 
coordinates { (0.7, 0.6775) +- (0, 0.4677) (0.8, 0.8187) +- (0, 0.3855) (0.85, 0.8875) +- (0, 0.3162) (0.9, 0.9175) +- (0, 0.2753) (0.95, 0.9637) +- (0, 0.1870) (0.98, 0.9788) +- (0, 0.1443) (0.99, 1.0000) +- (0, 0.0) };
\addlegendentry{\texttt{gtz} (cp)}

\addplot[color=cbblue, solid, mark=square*, error bars/.cd, y dir=both, y explicit, error bar style={opacity=0.4, thin}] 
coordinates { (0.7, 0.6998) +- (0, 0.4584) (0.8, 0.8025) +- (0, 0.3981) (0.85, 0.8544) +- (0, 0.3527) (0.9, 0.9022) +- (0, 0.2970) (0.95, 0.9514) +- (0, 0.2150) (0.98, 0.9804) +- (0, 0.1387) (0.99, 0.9899) +- (0, 0.1001) };
\addlegendentry{\texttt{dbp} (hcc)}

\addplot[color=cbblue, dashed, mark=square, error bars/.cd, y dir=both, y explicit, error bar style={opacity=0.4, thin}] 
coordinates { (0.7, 0.6998) +- (0, 0.4584) (0.8, 0.8025) +- (0, 0.3981) (0.85, 0.8544) +- (0, 0.3527) (0.9, 0.9022) +- (0, 0.2970) (0.95, 0.9514) +- (0, 0.2150) (0.98, 0.9804) +- (0, 0.1387) (0.99, 0.9899) +- (0, 0.1001) };
\addlegendentry{\texttt{dbp} (cp)}

\addplot[color=cbgreen, solid, mark=triangle*, error bars/.cd, y dir=both, y explicit, error bar style={opacity=0.4, thin}] 
coordinates { (0.7, 0.6977) +- (0, 0.4592) (0.8, 0.7986) +- (0, 0.4010) (0.85, 0.8541) +- (0, 0.3530) (0.9, 0.9083) +- (0, 0.2886) (0.95, 0.9598) +- (0, 0.1964) (0.98, 0.9882) +- (0, 0.1081) (0.99, 0.9950) +- (0, 0.0702) };
\addlegendentry{\texttt{img} (hcc)}

\addplot[color=cbgreen, dashed, mark=triangle, error bars/.cd, y dir=both, y explicit, error bar style={opacity=0.4, thin}] 
coordinates { (0.7, 0.6977) +- (0, 0.4592) (0.8, 0.7967) +- (0, 0.4024) (0.85, 0.8469) +- (0, 0.3601) (0.9, 0.8982) +- (0, 0.3025) (0.95, 0.9508) +- (0, 0.2163) (0.98, 0.9807) +- (0, 0.1375) (0.99, 0.9911) +- (0, 0.0938) };
\addlegendentry{\texttt{img} (cp)}

\end{axis}
\end{tikzpicture}
\begin{tikzpicture}
\begin{axis}[
    title={PS Size vs $1-\alpha$},
    xlabel={$1-\alpha$},
    ylabel={PS Size},
    xmin=0.65, xmax=1.0,
    ymode=log,
    xtick={0.7, 0.8, 0.85, 0.9, 0.95, 0.98, 0.99},
    xticklabels={.7, .8, .85, .9, .95, .98, .99},
    xticklabel style={
        yshift={mod(\ticknum,2) == 0 ? "0pt" : "-10pt"},
        font=\small
    },
    legend pos=north west,
    legend columns=2,
    legend style={
        nodes={scale=0.6, transform shape}
    },
    grid=major,
    width=.32\textwidth,
    height=5.5cm
]

\addplot[color=cbvermillion, solid, mark=*, error bars/.cd, y dir=both, y explicit, error bar style={opacity=0.4, thin}] 
coordinates { (0.7, 1.19) +- (0, 0.39) (0.8, 1.22) +- (0, 0.42) (0.85, 1.23) +- (0, 0.42) (0.9, 1.19) +- (0, 0.39) (0.95, 1.14) +- (0, 0.34) (0.98, 1.07) +- (0, 0.26) (0.99, 1.05) +- (0, 0.22) };
\addlegendentry{\texttt{gtz} HCC}

\addplot[color=cbvermillion, dashed, mark=o, error bars/.cd, y dir=both, y explicit, error bar style={opacity=0.4, thin}] 
coordinates { (0.7, 1.27) +- (0, 0.47) (0.8, 2.12) +- (0, 1.17) (0.85, 2.73) +- (0, 1.59) (0.9, 3.25) +- (0, 1.87) (0.95, 4.26) +- (0, 2.25) (0.98, 5.12) +- (0, 2.47) (0.99, 8.34) +- (0, 2.44) };
\addlegendentry{\texttt{gtz} CP}

\addplot[color=cbblue, solid, mark=square*, error bars/.cd, y dir=both, y explicit, error bar style={opacity=0.4, thin}] 
coordinates { (0.7, 0.7) +- (0, 0.46) (0.8, 0.8) +- (0, 0.4) (0.85, 0.86) +- (0, 0.35) (0.9, 0.9) +- (0, 0.3) (0.95, 0.95) +- (0, 0.21) (0.98, 0.98) +- (0, 0.13) (0.99, 1.0) +- (0, 0.07) };
\addlegendentry{\texttt{dbp} HCC}

\addplot[color=cbblue, dashed, mark=square, error bars/.cd, y dir=both, y explicit, error bar style={opacity=0.4, thin}] 
coordinates { (0.7, 0.7) +- (0, 0.46) (0.8, 0.8) +- (0, 0.4) (0.85, 0.86) +- (0, 0.35) (0.9, 0.9) +- (0, 0.3) (0.95, 0.95) +- (0, 0.21) (0.98, 0.98) +- (0, 0.13) (0.99, 1.0) +- (0, 0.07) };
\addlegendentry{\texttt{dbp} CP}

\addplot[color=cbgreen, solid, mark=triangle*, error bars/.cd, y dir=both, y explicit, error bar style={opacity=0.4, thin}] 
coordinates { (0.7, 0.79) +- (0, 0.41) (0.8, 0.98) +- (0, 0.27) (0.85, 1.13) +- (0, 0.39) (0.9, 1.43) +- (0, 0.90) (0.95, 2.56) +- (0, 3.36) (0.98, 3.73) +- (0, 5.18) (0.99, 4.32) +- (0, 5.83) };
\addlegendentry{\texttt{img} HCC}

\addplot[color=cbgreen, dashed, mark=triangle, error bars/.cd, y dir=both, y explicit, error bar style={opacity=0.4, thin}] 
coordinates { (0.7, 0.79) +- (0, 0.41) (0.8, 1.02) +- (0, 0.33) (0.85, 1.24) +- (0, 0.51) (0.9, 1.75) +- (0, 1.15) (0.95, 3.65) +- (0, 4.33) (0.98, 11.04) +- (0, 19.7) (0.99, 27.83) +- (0, 55.02) };
\addlegendentry{\texttt{img} CP}
\end{axis}
\end{tikzpicture}
\begin{tikzpicture}
\begin{axis}[
    title={Covered leaves vs $1-\alpha$},
    xlabel={$1-\alpha$},
    ylabel={Covered leaves},
    xmin=0.65, xmax=1.0,
    ymode=log,
    legend columns=2,
    legend style={
        nodes={scale=0.6, transform shape}
    },
    xtick={0.7, 0.8, 0.85, 0.9, 0.95, 0.98, 0.99},
    xticklabels={.7, .8, .85, .9, .95, .98, .99},
    xticklabel style={
        yshift={mod(\ticknum,2) == 0 ? "0pt" : "-10pt"},
        font=\small
    },
    grid=major,
    width=.32\textwidth,
    height=5.5cm
]

\addplot[color=cbvermillion, solid, mark=*, error bars/.cd, y dir=both, y explicit, error bar style={opacity=0.4, thin}] 
coordinates { (0.7, 1.37) +- (0, 0.95) (0.8, 4.21) +- (0, 3.95) (0.85, 5.29) +- (0, 4.14) (0.9, 6.17) +- (0, 4.14) (0.95, 7.34) +- (0, 3.88) (0.98, 8.18) +- (0, 3.51) (0.99, 9.48) +- (0, 1.96) };
\addlegendentry{\texttt{gtz} HCC}

\addplot[color=cbvermillion, dashed, mark=o, error bars/.cd, y dir=both, y explicit, error bar style={opacity=0.4, thin}] 
coordinates { (0.7, 1.27) +- (0, 0.47) (0.8, 2.12) +- (0, 1.17) (0.85, 2.73) +- (0, 1.59) (0.9, 3.25) +- (0, 1.87) (0.95, 4.26) +- (0, 2.25) (0.98, 5.12) +- (0, 2.47) (0.99, 8.34) +- (0, 2.44) };
\addlegendentry{\texttt{gtz} CP}

\addplot[color=cbblue, solid, mark=square*, error bars/.cd, y dir=both, y explicit, error bar style={opacity=0.4, thin}] 
coordinates { (0.7, 0.7) +- (0, 0.46) (0.8, 0.8) +- (0, 0.4) (0.85, 0.86) +- (0, 0.35) (0.9, 0.9) +- (0, 0.3) (0.95, 0.95) +- (0, 0.21) (0.98, 0.98) +- (0, 0.13) (0.99, 1.0) +- (0, 0.07) };
\addlegendentry{\texttt{dbp} HCC}

\addplot[color=cbblue, dashed, mark=square, error bars/.cd, y dir=both, y explicit, error bar style={opacity=0.4, thin}] 
coordinates { (0.7, 0.7) +- (0, 0.46) (0.8, 0.8) +- (0, 0.4) (0.85, 0.86) +- (0, 0.35) (0.9, 0.9) +- (0, 0.3) (0.95, 0.95) +- (0, 0.21) (0.98, 0.98) +- (0, 0.13) (0.99, 1.0) +- (0, 0.07) };
\addlegendentry{\texttt{dbp} CP}

\addplot[color=cbgreen, solid, mark=triangle*, error bars/.cd, y dir=both, y explicit, error bar style={opacity=0.4, thin}] 
coordinates { (0.7, 0.79) +- (0, 0.41) (0.8, 1.24) +- (0, 1.91) (0.85, 2.06) +- (0, 3.74) (0.9, 3.87) +- (0, 7.30) (0.95, 12.50) +- (0, 35.90) (0.98, 85.85) +- (0, 221.61) (0.99, 186.55) +- (0, 337.20) };
\addlegendentry{\texttt{img} HCC}

\addplot[color=cbgreen, dashed, mark=triangle, error bars/.cd, y dir=both, y explicit, error bar style={opacity=0.4, thin}] 
coordinates { (0.7, 0.79) +- (0, 0.41) (0.8, 1.02) +- (0, 0.33) (0.85, 1.24) +- (0, 0.51) (0.9, 1.75) +- (0, 1.15) (0.95, 3.65) +- (0, 4.33) (0.98, 11.04) +- (0, 19.7) (0.99, 27.83) +- (0, 55.02) };
\addlegendentry{\texttt{img} CP}

\end{axis}
\end{tikzpicture}
\caption{Additional results for various desired coverage ratios $\alpha$.
HCC maintains coverage and significantly reduces the nominal prediction set size at higher $1-\alpha$, at a slight increase of \# covered leaves. Results for \texttt{dbp} overlap. Error bars indicate SD.}
\label{fig:additional_alpha}
\end{figure}
We list our main results in Table~\ref{tab:base_results}. We observe that the coverage requirement of $1-\alpha$ is met for most baselines and some ablations, and that HCC produces solutions with the lowest (\texttt{dbp}, \texttt{img}) or comparable to the lowest cost (\texttt{gtz}). Furthermore, we note in the column \emph{PS size}, which denotes the nominal efficiency defined as the average nominal number of elements in the prediction set, that HCC produces substantially smaller prediction sets across data sets. Next, in the column \emph{covered leaves} we see that the nominally smaller prediction sets come at a moderate cost to the total number of covered leaves i.e. the leaf efficiency.
Furthermore, Figure~\ref{fig:betas} shows how HCC allows a fine-grained control of the prediction set size against the number of covered leaves across $\beta$ values. The initial search spaces were constructed in .28, .02, .54 seconds for \texttt{img}, \texttt{gtz}, and \texttt{dbp}, and the average number of candidates after dynamic pruning was 15, 10, 10 respectively. See Appendix~\ref{app:sample_outputs}.s

\paragraph{User Study}
We also assess the utility of HCC with a preliminary user study, in which the authors provided their preferences for HCC or standard CP in 500 random samples of the \texttt{dbp} and \texttt{img} datasets. The inter-annotator agreement was fair to moderate with Fleiss's $\kappa$ of .47 for \texttt{dbp} and .53 for \texttt{img}. A binomial test indicates a significant preference for HCC over standard CP in 57\% and 71\% of cases, respectively, for \texttt{dbp} and \texttt{img}, highlighting HCC's practical utility (see Appendix~\ref{app:user_study}).

\section{Discussion}
\label{sec:discussion}
This work is, to the best of our knowledge, the first to propose an efficient search for the problem of conformal prediction with marginal coverage guarantees over a hierarchical label space that allows explicitly balancing nominal prediction set size with specificity, which is here quantified as the number of covered leaves.
 While our results are promising, we believe that there is room to improve and extend our approach. For example, our preliminary user study encourages a larger user study on a realistic use case involving e.g. intent recognition or diagnosis~\citep{hengst-etal-2024-conformal}. Here, we can also consider optimizing $\beta$ adaptively based on individual user preferences since the trade-off between specificity and compactness may be relative to the user and domain. 

Additionally, our approach core can be improved in various ways. Firstly, we apply a correction approach inspired by Westfall and Young's MaxT during inference. While this correction is computationally efficient due to using a single step and only requiring the empirical distribution of nonconformity scores already computed during calibration, this mechanism may be further improved using a multi-step correction mechanism. This would require ordering the calibrated predictors in $\mathcal{H}$ meaningfully a priori based on the expected costs associated with each predictor in $\mathcal{H}$, which may be empirically more preformant but will, the best of our understanding, require additional assumptions on the DAG in order to maintain guarantees.

Moreover, we believe that there is room for computational improvement by further pruning the search space, potentially in ways that would also reduce the correction to $\alpha$ and therefore is expected to further improve performance. There are various directions here, including using the hierarchy to assess which conformal predictors are likely to produce similar sets, and pruning predictors dynamically based on the empirical distribution of scores over all DAG nodes in the calibration set.

\paragraph{Limitations}%
 Our approach produces modest worst-case gains in search space pruning down to  $\mathcal{O}(1.5028^{2^d})$ for a hierarchy of depth $d$, which may be intractable for large $d$. In these cases, a per-level fallback can be used that maintains coverage guaranttees, but lacks optimality. The user study suffers from bias as the authors served as annotators, and blinding was not practically achievable due to smaller nominal set sizes for HCC over the alternative. Finally, we set $\beta$ heuristically for the main results, while it might depend on the user and domain. Future work should investigate adaptively tuning $\beta$ e.g. based on interactions with users.

\section{Conclusion}%
\label{sec:conclusion}
In this paper, we argued the importance of background knowledge in creating informative, concise, and connected conformal prediction sets. We formally introduced the task of hierarchical conformal classification, in which the target labels for classification are organized in a hierarchy. We proposed a solution based on a search over the space of relevant conformal predictors for the given hierarchy and evaluate it quantitatively using datasets with various modalities, and in a user study.
Our results indicate that hierarchical information can lead to better prediction sets in both types of evaluation, and we find that annotators significantly prefer our approach over a non-hierarchical output.

\section*{\normalsize \textbf{Acknowledgments \& Funding Statement}}
We kindly thank our reviewers for their time and their useful comments.
We thank Xander Wilcke for the useful discussions during the initial phases of this project. This work was partially supported by the Hybrid Intelligence Centre of the Dutch Ministry of Education, Culture and Science through the Netherlands Organisation for Scientific Research (NWO), grant number 024.004.022. All authors declare not competing interests.

\bibliographystyle{apalike}
\bibliography{bibliography}

\clearpage
\appendix
\onecolumn

\section{Proofs}
\label{app:proofs}
We here repeat and provide proofs for all theoretical results.

\PropCoverage*
\begin{proof}
Suppose, for contradiction, that there exists a subset \( \mathcal{M} \subseteq \mathcal{V} \) such that \( \text{leaf-cov}_{\mathcal{T}}(\mathcal{M}) \subset \mathcal{Y} \). Then, there exists at least one leaf \( y^* \in \mathcal{Y} \) such that \( y^* \notin \text{leaf-cov}_{\mathcal{T}}(\mathcal{M}) \). 

Since \( y^* \notin \text{leaf-cov}_{\mathcal{T}}(\mathcal{M}) \), we have \( \mathbb{P}(y^* \in \text{leaf-cov}_{\mathcal{T}}(\mathcal{M})) = 0 \). Therefore,
\[
\mathbb{P}(y^* \in \text{leaf-cov}_{\mathcal{T}}(\mathcal{M})) \leq 1 - \mathbb{P}(Y = y^*).
\]
As a result, for any \( \alpha \in [0, \mathbb{P}(Y = y^*)) \), we have
\[
\mathbb{P}(y^* \in \text{leaf-cov}_{\mathcal{T}}(\mathcal{M})) < 1 - \alpha
\]
which contradicts with the requirement in equation~\eqref{eq:objective}.
\end{proof}

\vspace{0.1cm}

\PropAncestral*
\begin{proof}
Let \( \mathcal{N}^* \subseteq \mathcal{V} \) be any feasible solution to the objective in equation~\eqref{eq:objective} and let $\mathcal{S}_{\text{NOL}}$ denote the set of all non-overlapping leaf covers following Definition~\ref{def:nol-cover}.
If \( \mathcal{N}^* \in \mathcal{S}_{\mathrm{NOL}} \), we are done. Otherwise, we show how to construct a set \( \mathcal{S} \in \mathcal{S}_{\mathrm{NOL}} \) with lower or equal cost.

We construct \( \mathcal{S} \) from \( \mathcal{N}^* \) by removing any node \( u \in \mathcal{N}^* \) that is a proper ancestor of another node \( v \in \mathcal{N}^* \) iteratively until no more ancestors are present.

The resulting set \( \mathcal{S} \subseteq \mathcal{V} \) now satisfies by construction:
    \[
    \forall u, v \in \mathcal{S},\ u \neq v \Rightarrow \neg(u \precprec v) \land \neg(v \precprec u).
    \]
and both \( |\mathcal{S}| \leq |\mathcal{N}^*| \) and \( \left| \bigcup_{v \in \mathcal{S}} \text{leaf-cov}_\mathcal{T}(v) \right| \leq \left| \bigcup_{v \in \mathcal{N}^*} \text{leaf-cov}_\mathcal{T}(v) \right| \), since redundant coverage was eliminated.

Therefore, \( \mathcal{S} \in \mathcal{S}_{\mathrm{NOL}} \) is a feasible solution with objective cost no greater than that of \( \mathcal{N}^* \). This holds for all feasible \( \mathcal{N}^* \), so the search can be restricted to the conformal classifiers calibrated on the NOL-cover set without incurring additional cost.
\end{proof}

\vspace{0.1cm}

\CoverageTheorem*
\begin{proof}
By construction, the score \( s(\mathbf{x}_i, \mathbf{y}_i, \mathcal{S}) \) selects the node \( v' \in \mathcal{S} \cap \mathcal{Y}' \) in the NOL-cover \(\mathcal{S}\) with the highest classification score and assigns the score \( \hat{g}(\mathbf{x})_{v'} \). Let \(s_i\) denote this score for some particular pair \((\mathbf{x}_i, \mathbf{y}_i)\) and an arbitrary NOL-cover $\mathcal{S}$.

Given the exchangeability of the calibration examples and the test instance, the set of scores \( \{ s_1, \dots, s_{n_c}, s_{t}\} \) is also exchangeable. Therefore, the rank of \( s_{t} \) among these \( n_c + 1 \) scores is uniformly distributed over \( \{1, \dots, n_c + 1\} \). 

It follows that:
\[
\mathbb{P}\left[ s_{t} \leq \hat{q} \right] \geq 1 - \alpha.
\]

Now observe that:
\[
s_{t} \leq \hat{q} \iff \hat{g}(\mathbf{x}_{t})_{v'_{t}} \leq \hat{q} \iff v'_{t} \in C(\mathbf{x}_{t}),
\]
where \( v'_{t} := \arg\max_{j \in \mathcal{S} \cap \mathcal{Y}_{t}'} \hat{g}(\mathbf{x}_{t})_j \) is a node in the NOL-cover which has been labeled as ground truth.

By definition of the propagated label set, since \( y_{t} \in \text{leaf-cov}_{\mathcal{T}}(\{v'_{t}\}) \), and \( v'_{t} \in C(\mathbf{x}_{t}) \), it follows that:
\[
y_{t} \in \text{leaf-cov}_{\mathcal{T}}\bigl(C(\mathbf{x}_{t})\bigr).
\]

Therefore,
\[
\mathbb{P} \left[ y_{t} \in \text{leaf-cov}_{\mathcal{T}}\bigl(C(\mathbf{x}_{t})\bigr) \right] = \mathbb{P} \left[ s_{t} \leq \hat{q} \right] \geq 1 - \alpha.
\]
\end{proof}

\PropPrune*
\begin{proof}
Let $\mathcal{T} = (\mathcal{V},\prec)$ be a hierarchy with leaves $\mathcal{Y}$, and $\precprec$ its strict ancestor relation as before. Now let $h_{\beta,\mathcal{S}}^*$ be an optimal predictor feasible to~\eqref{eq:objective} and assume it contains such an ancestor $v\in \mathcal{S}, v\precprec lca, lca \in LCA(C(\mathbf{x}_t,\mathcal{Y}))$ where $C(\mathbf{x}_t,\mathcal{Y})$ denotes the prediction set constructed according to Theorem~\ref{thm:coverage} over leaves $\mathcal{Y}$.

By construction
\begin{equation}
\label{eq:lca_size}
|\{lca\}| = 1 \leq |C(\mathbf{x}_t,\mathcal{S})|.
\end{equation}
Since $v \precprec lca$ and $lca \in LCA(C(\mathbf{x}_t,\mathcal{Y}))$,
\begin{align}
|\text{leaf-cov}_\mathcal{T}(\{lca\})| &\leq |\text{leaf-cov}_\mathcal{T}(v) |
\label{eq:leaf_cov}
\end{align}
Combining~\eqref{eq:lca_size} and~\eqref{eq:leaf_cov} shows $\mathcal{S}$ can be replaced with $\{lca\} \in LCA(C(\mathbf{x}_t,\mathcal{Y}))$ without risking optimality with respect to~\eqref{eq:objective}.
\end{proof}

\section{Sample outputs} 
\label{app:sample_outputs}
\begin{figure*}[h!]
  \centering
  \begin{subfigure}[b]{0.32\textwidth}
    \includegraphics[width=\linewidth]{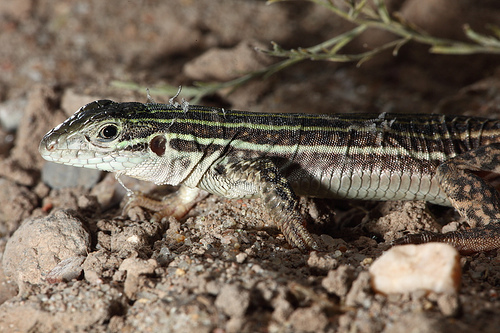}
    \caption{Users unanimously preferred HCC for this input with ground truth \texttt{whiptail}, HCC prediction  \texttt{\{lizard\}} and standard CP prediction \texttt{\{agama, alligator lizard, green lizard, whiptail\}}.}
    \label{fig:sub1}
  \end{subfigure}
  \hfill
  \begin{subfigure}[b]{0.32\textwidth}
    \includegraphics[width=\linewidth]{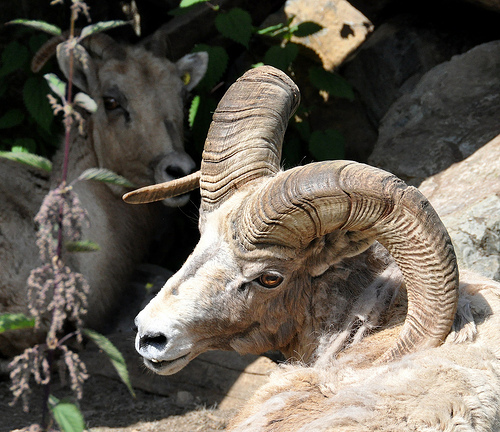}
    \caption{Users unanimously preferred standard CP for this input with ground truth \texttt{ram}, HCC prediction  \texttt{\{bovid\}} and standard CP prediction \texttt{\{ram, big horn\}}.}
    \label{fig:sub2}
  \end{subfigure}
  \hfill
  \begin{subfigure}[b]{0.32\textwidth}
    \includegraphics[width=\linewidth]{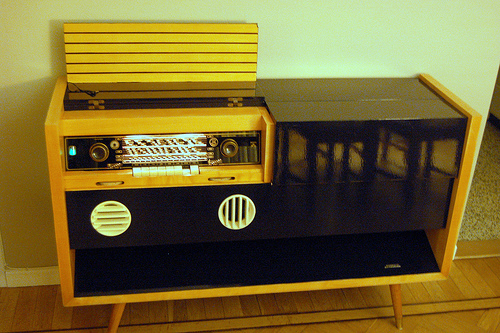}
    \caption{Users unanimously indicated no preference for this input with ground truth \texttt{tape player}, HCC prediction \texttt{\{whole\}} and standard CP prediction {\tiny\texttt{\{ashcan, ballpoint, bannister, bookcase, carpenter's kit, carton, cash machine, cassette, cassette player, cd player, chest, chiffonier, china cabinet, combination lock, computer keyboard, crash helmet, crate, desk, desktop computer, digital clock, dishwasher, drumstick, entertainment center, envelope, file, fire screen, fountain pen, freight car, garbage truck, grille, hammer, hard disc, harmonica, home theater, hook, joystick, laptop, loudspeaker, mailbox, marimba, microwave, modem, monitor, notebook, organ, paintbrush, paper towel, pay-phone, pencil sharpener, photocopier, pickup, plate rack, polaroid camera, pool table, printer, projector, radio, recreational vehicle, refrigerator, rubber eraser, safe, scale, school bus, screwdriver, slide rule, slot, space heater, spatula, stove, tape player, television, theater curtain, toaster, tobacco shop, toilet tissue, turnstile, upright, vending machine, volleyball, wallet\}}}.}
    \label{fig:sub3}
  \end{subfigure}
  \caption{Example output and user preferences for random samples from \texttt{img} with unanimous user preferences.}
  \label{fig:main}
\end{figure*}
\clearpage

\section{Worst Case Search Space Analysis}
\label{app:search_space_analysis}
The perfect binary tree is a balanced binary tree. For any such tree of depth $d$, the total number of nodes is
\[
|\mathcal{V}| = 2^{d+1} -1,
\]
and hence the total naive search space size is
\[
2^{|\mathcal{V}|} = 2^{2^{d+1}-1} = 2^-1 \cdot (2^2)^{2d} = \frac{1}{2} \cdot 4^{2^d}.
\]

The NOL-cover size can be inferred by defining the size recursively from the root down using the sequence \( C(d) \) as:
\[
\begin{aligned}
C(0) &= 1 \\
C(d) &= C(d-1)^2 + 1 \text{~~for~} d > 0
\end{aligned}
\]
since we need to include the left-hand side subtree and the right-hand side subtree (both at $C(d-1)$) together, or these subtree's ancestor at depth $d$ (hence $+1$).

The resulting sequence $(1, 5, 26, 677)$ is known as OEIS A003095 and can be approximated with the Aho-Sloane constant as $C(d) \approx 1.5028^{2^d}$~\cite{aho1973some}.

\section{Search Space Construction Algorithm}
An algorithm for finding NOL-covers is shown in Alg.~\ref{alg:nol-covers}. It generates all valid NOL-covers in a top-down recursive way, starting from the set of root nodes in \(\mathcal{T}\) (ln 16-18), which is a NOL-cover by definition. The recurrence step involves first validating that the current candidate contains no overlapping nodes and has not already been added to the result (ln 2-4). The candidate is added to the result if it covers all leaves (ln 5-6). New candidates are formed by replacing each node in the set with its children, and these new candidates are then recurred upon (ln 7-12) until all NOL-covers are found.
\begin{algorithm}[h]
    \begin{flushleft}
\caption{Algorithm for finding all NOL-covers}
\label{alg:nol-covers}
\textbf{Input}: DAG \(\mathcal{T}\) with nodes $\mathcal{V}$ \\
\textbf{Output}: Non-overlapping leaf covers
\end{flushleft}
\begin{algorithmic}[1]
\PROCEDURE{\textsc{recur}}{candidate, \(\mathcal{T}\), result}
    \IF{\textsc{ancestors}(candidate, \(\mathcal{T}\)) \OR candidate $\in$ result}
        \STATE \textbf{return}                                         \hfill \COMMENT{invalid candidate}
    \ENDIF
    \IF{cov$_T$(candidate) $=$ leaves(\(\mathcal{T}\))}                 
        \STATE result.append(candidate)                     \hfill \COMMENT{add candidates}
        \FORALL{n $\in$ candidate}
            \STATE new-candidate $\gets$ candidate.copy()           
            \STATE new-candidate.remove(n)                   \hfill \COMMENT{replace by children}
            \STATE new-candidate.update(n.expand())         
            \STATE \textbf{call} \textsc{recur}(new-candidate, \(\mathcal{T}\), result)
        \ENDFOR
    \ENDIF
    \STATE \RETURN
\ENDPROCEDURE

\STATE nol-covers $\gets$ List()                     
\STATE candidate $\gets$ Set(root(\(\mathcal{T}\)))              \hfill \COMMENT{initial candidate}   
\STATE \textsc{recur}(candidate, \(\mathcal{T}\), nol-covers)     \hfill \COMMENT{start recursion}
\STATE \RETURN ~nol-covers
\end{algorithmic}

\end{algorithm}

While Alg.~\ref{alg:nol-covers} needs to be run only once per taxonomy and can in practice be implemented efficiently using caching and parallel execution, some taxonomies may generate a large search space, resulting in long running times and large corrections to the allowed error $\alpha$. In this case, a limited search can be employed, based on the depths of taxonomy \(\mathcal{T}\) with maximum depth $d$, where a NOL-cover for each depth $0,1,\dots,d_i,\dots,d$ contains all nodes at that depth $d_i$ and all leaves at a lower depth $d_j \leq d_i$.
In this case, an approximation of~\eqref{eq:objective} is obtained by searching in a space consisting of $d$ conformal predictors.

\FloatBarrier

\section{Data and Taxonomies}
We here list the distribution of nodes per level for all taxonomies and visualise the taxonomies for \texttt{dbp} and \texttt{gtz}. The full taxonomy for \texttt{img} is omitted due to size.

\begin{figure}[h!] 
\centering
\includegraphics[trim={0.7cm 2.1cm 0cm 2cm}, clip, width=\textwidth]{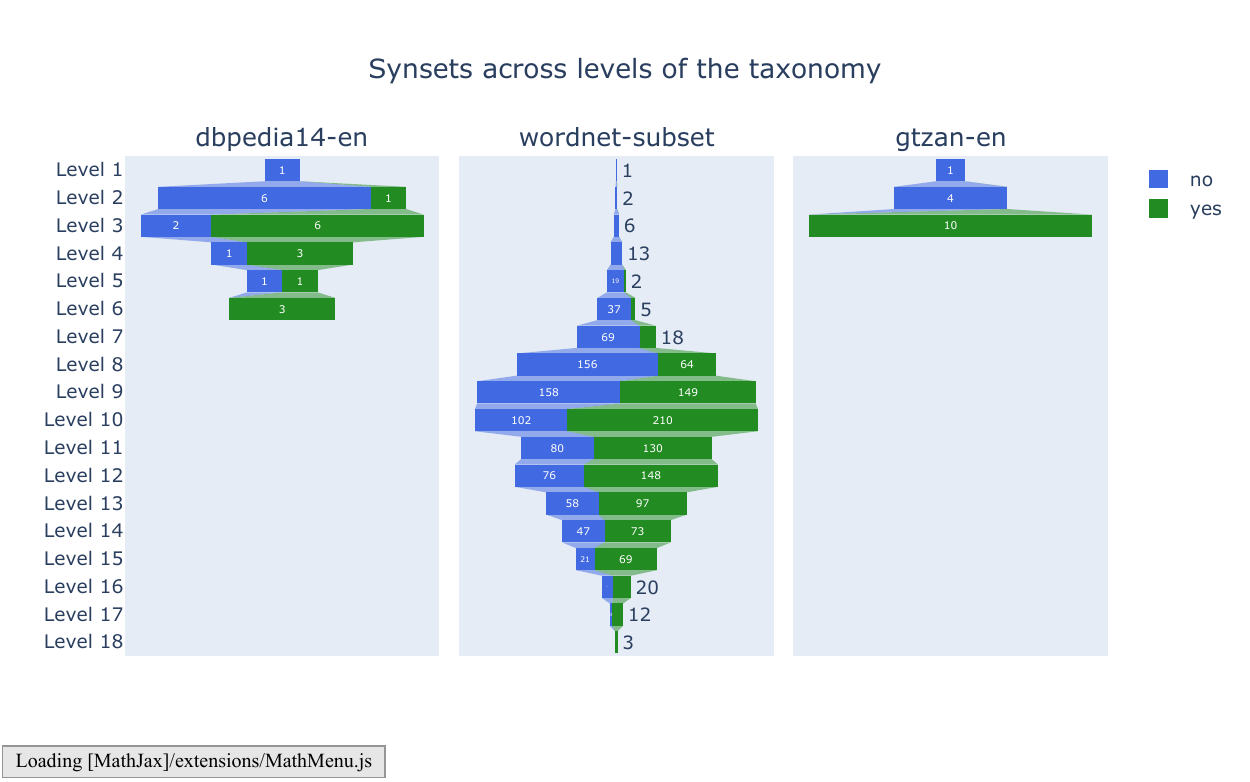}
    \caption{Number of nodes per level: \texttt{dbp} (left) and \texttt{wordnet-subset} (center), \texttt{gtz} (right). "yes" denotes the number of leaves, and "no" the number of internal nodes per level.}
    \label{fig:funnel-taxonomy}
\end{figure}

\begin{figure*}[h!]
\centering
\includegraphics[trim={0.2cm 10cm 2.5cm .1cm}, clip, width=\textwidth]{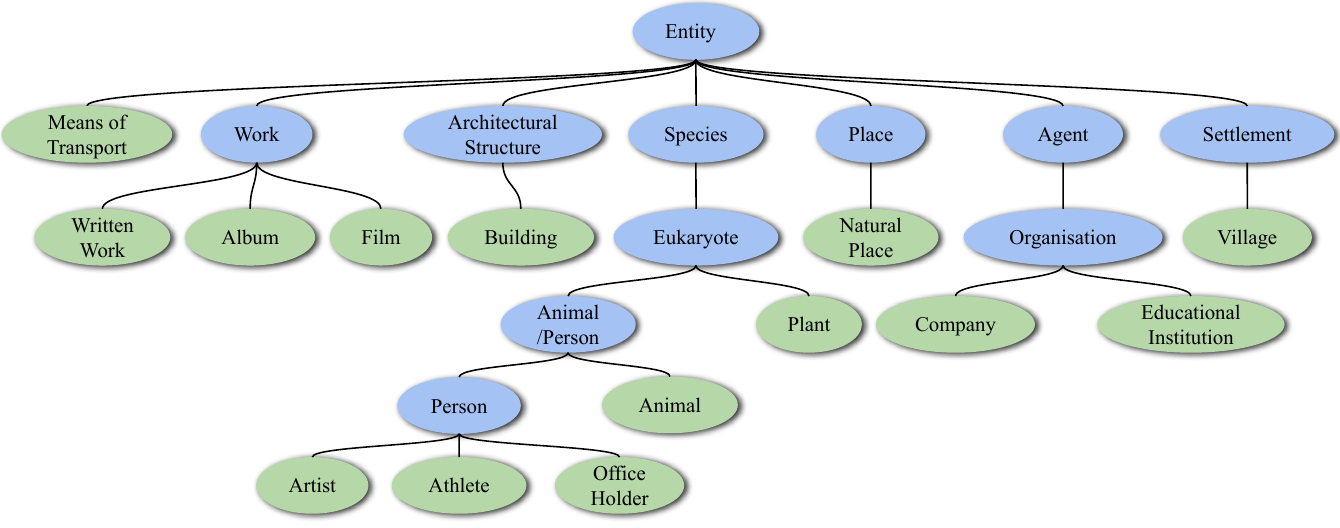}
\caption{\texttt{dbp} taxonomy. Leaves are denoted in green and intermediate nodes in blue.}
\label{fig:dbpedia14-vis}
\end{figure*}

\begin{figure*}[h!]
\centering
\includegraphics[trim={0.1cm 14.5cm 3.3cm .1cm}, clip, width=\textwidth]{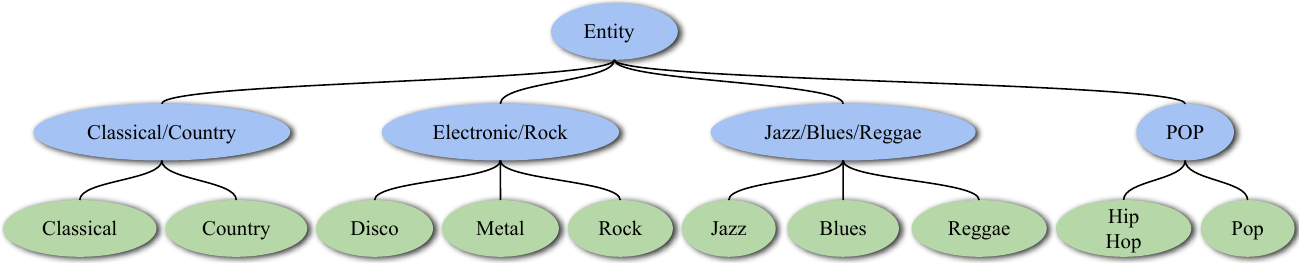}
\caption{\texttt{gtz} taxonomy.}
\label{fig:gtzan-vis}
\end{figure*}

\FloatBarrier
\section{Additional Results \& Ablation Details}
Here we provide the full result set for CSP, comparing it with our approach and baselines. We selected values  for hyperparameter $M$ based on the author's suggestions in~\cite{zhang2024conformal}, the main results in Table~\ref{tab:base_results} have been selected based on $M$ values with minimal costs.

\begin{table*}[h!]
\centering\caption{Additional test set results for CSP.}
\label{tab:additional_results}
\setlength{\tabcolsep}{3.5pt}
\begin{tabular}{l l l l r@{\hskip 2pt}l r@{\hskip 2pt}l r@{\hskip 2pt}l r@{\hskip 2pt}l r@{\hskip 2pt}l}\toprule
     & \(\beta\) & \(1-\alpha\) & method           & \multicolumn{2}{c}{coverage $\uparrow$} 
                     & \multicolumn{2}{c}{costs $\downarrow$} 
                     & \multicolumn{2}{c}{PS size $\downarrow$} 
                     & \multicolumn{2}{c}{covered leaves $\downarrow$} \\
\midrule
 \texttt{dbp} & .19   & .9997 & Standard CP                    & \textbf{.9998} & $\pm$ .02 &      3.18 & $\pm$ 2.62  &       2.67 & $\pm$ 2.20  &      \textbf{2.67} & $\pm$ 2.20     \\
   & &  & LCA                        & \textbf{.9998} & $\pm$ .02 &      2.2  & $\pm$1.13   &       \textbf{1}    & $\pm$ .00  &      6.32 & $\pm$ 5.94     \\
   & &  & CSP $M=4$                   & .9938 & $\pm$ .08 &      1.2  & $\pm$ .15  &       1.01 & $\pm$ .09  &      1.02 & $\pm$ .50     \\
   & &  & CSP $M=8$                   & .9938 & $\pm$ .08 &      1.2  & $\pm$ .15  &       1.01 & $\pm$ .09  &      1.02 & $\pm$ .50     \\
   & &  & HCC (ours)              & \textbf{.9999} & $\pm$ .01 &      \textbf{2.01} & $\pm$ .94  &       1.23 & $\pm$ .45  &      4.12 & $\pm$ 4.47     \\
\texttt{img} & .04 & .98  & Standard CP                     & \textbf{.9807} & $\pm$ .14 &     11.49 & $\pm$ 20.49 &      11.04 & $\pm$ 19.70 &     \textbf{11.04} & $\pm$ 19.70    \\
  & &   & LCA                        & \textbf{.9807} & $\pm$ .14 &     14.79 & $\pm$15.63  &       \textbf{1}    & $\pm$ .00  &    344.69 & $\pm$ 390.64   \\
  & &   & CSP $M=4$                   & \textbf{.9900}   & $\pm$ .10 &     17.88 & $\pm$ 16.93 &       3.01 & $\pm$ 1.22  &    368.36 & $\pm$ 400.30   \\
  & &   & CSP $M=8$                   & \textbf{.9850}  & $\pm$ .12 &     15.98 & $\pm$ 17.03 &       5.03 & $\pm$ 2.98  &    269.83 & $\pm$ 364.12   \\
  & &   & HCC (ours)              & \textbf{.9882} & $\pm$ .11 &      \textbf{7.17} & $\pm$ 10.26 &       3.73 & $\pm$ 5.18  &     85.85 & $\pm$ 221.61   \\
  \texttt{gtz} & .25 & .9 & Standard CP & \textbf{0.9175} & $\pm 0.28$ &      4.07 & $\pm 2.34$  &       3.25 & $\pm$ 1.87 &      \textbf{3.25} & $\pm$ 1.87     \\
  & &   & LCA    & \textbf{0.9175} & $\pm$ 0.28 &      2.87 & $\pm$1.0     &       \textbf{1}    & $\pm$ .00 &      7.47 & $\pm$ 4.01     \\
  & &   & CSP $M=4$                   & \textbf{.9762} & $\pm$ .15 &      4.67 & $\pm$ 1.67  &       3.3  & $\pm$ 1.09  &      5.53 & $\pm$ 2.73     \\
  & &   & CSP $M=8$                   & \textbf{.9712} & $\pm$ .17 &      5.32 & $\pm$ 2.23  &       4.05 & $\pm$ 1.71  &      5.08 & $\pm$ 2.30     \\
  & &   & HCC (ours)              & \textbf{.9275} & $\pm$ .26 &      \textbf{2.47} & $\pm$ .94  &       1.32 & $\pm$ .47  &      4.61 & $\pm$ 3.80     \\
\bottomrule
\end{tabular}
\end{table*}

Additionally, we provide runtime results in a non-cached and a cached inference regime. In the latter, calibrators, standard CP and LCA results are cached and only optimization for a different $\beta$ is performed for each test instance.

\begin{table}[h!]
\centering
\caption{Runtime in seconds of HCC per stage, with inference served from the disk cache,
for the main results in Table~\ref{tab:base_results}. Totals exclude data loading. For reference,
we list the HCC total without cache, i.e.\ including per-instance inference over all candidate
covers, and the total runtimes of Standard CP and CSP ($M=8$). All runs are single-process
on the hardware described in Appendix~\ref{app:compute}.}
\label{tab:runtime}
\small
\begin{tabular}{lrrr}
\toprule
 & \texttt{dbp} & \texttt{img} & \texttt{gtz} \\
\midrule
Search space construction (NOL-covers)  &    0.28 &    0.54 &  0.02 \\
Score propagation                       &    0.23 &   15.88 & <0.01 \\
Calibration                             &    9.00 &   19.36 &  0.03 \\
Cache lookup (key and load)             &    5.83 &    8.11 &  0.03 \\
Prediction set selection                &    8.53 &   49.95 &  0.01 \\
\midrule
HCC total, cached inference             & \textbf{23.87} & \textbf{93.84} & \textbf{0.09} \\
HCC total, without cache                & 2,832.21 &  986.88 &  3.82 \\[.3ex]
Standard CP                             &   25.3  &    1.42 &  0.1  \\
CSP ($M=8$)                             & 114,475.3 & 81,768.0 & 136.9 \\
\bottomrule
\end{tabular}
\end{table}

\begin{table*}
\end{table*}

\section{User Study Details}
\label{app:user_study}
We assessed our approach using a preliminary user study, in which the authors provided their preference on 500 random samples with differing prediction sets for the \texttt{dbp} and \texttt{img} datasets. Annotators were asked to indicate their preference for one result over the other or to indicate no preference. The Fleiss' Kappa values for inter-annotator agreement were .47 and .53 for these datasets, respectively. This indicates a fair to moderate level of agreement. 


Table~\ref{table:user-study-results} shows the annotation and majority distribution across the labels \emph{HCC}, \emph{standard CP}, and \emph{no preference}, for both \texttt{dbp} and \texttt{img.} Overall, HCC was favoured by annotators, both in annotation numbers (52.4\% and 57.9\% for \texttt{dbp} and \texttt{img}) and in majority distribution numbers (50.4\% for both).

We applied the binomial test to evaluate whether the number of annotators who preferred hierarchical over Standard CP differed significantly from what would be expected by chance ($p=0.05$). The binomial tests indicate a significant preference for hierarchical over Standard CP for both datasets, with test statistics of .57 for \texttt{dbp} and .71 for \texttt{img}, and p-values of \num{3.4e-05} and \num{2.4e-33}. This means that in approximately 57\% and 71\% of cases, respectively, annotators preferred hierarchical over Standard CP, which is significantly higher than expected by chance.

User preferences varied by domain and task. In \texttt{dbp}, differing taxonomies and HCC's abstract outputs made prediction sets less helpful, while standard CP was more precise at the cost of larger sets. In \texttt{img}, users preferred standard CP for small sets and HCC when unsure or faced with fine-grained categories. Abstract labels (e.g., ``object'') were generally unhelpful. Preferences often depended on whether the task called for broad recognition or fine-grained detail, suggesting future evaluations should consider task context. Moreover, in \texttt{img}, several samples consist of multiple objects in one image, while the ground truth label is only one, i.e., each item has a single label rather than multiple labels, and the classifier is trained on multiclass classification rather than multilabel classification. In such cases, standard CP often predicts a large set of labels while HCC may prefer ancestors, which can usually lead to abstract outputs such as "object" or "artifact". An in-depth study of the impact of such samples on different CP models can be a direction for future research. 

\begin{table}[h]

\centering
\caption{ Annotation and majority distribution across the preferences \emph{HCC} (ours), \emph{Standard CP} and \emph{No Preference}.}
\label{table:user-study-results}
\small
\begin{tabular}{llrrr}
\toprule
&\multirow{2}{*}{Metric} & \multicolumn{3}{c}{Preference} \\
& & \multicolumn{1}{c}{HCC} & \multicolumn{1}{c}{Standard CP} & \multicolumn{1}{c}{No Preference} \\
\midrule
 \texttt{dbp} & Annotation distribution (\#) &  523  & 401  & 74  \\
 &Annotation distribution (\%) & 52.4 & 40.2 & 7.4 \\
 &Majority distribution (\#) & 126 & 90 & 12 \\
 &Majority distribution (\%) & 50.4 & 36.0 & 4.8     \\
 \midrule
 \texttt{img} & Annotation distribution (\#) &  577  & 239  & 180  \\
 &Annotation distribution (\%) &  57.9 & 24.0 & 18.1 \\
 &Majority distribution (\#) & 126 & 35 & 46 \\
 &Majority distribution (\%) & 50.4 & 14.0 & 18.4 \\
\bottomrule
\end{tabular}

\end{table}

\section{Compute Infrastructure} 
\label{app:compute}
All experiments were conducted on a machine equipped with an Apple M2 Pro chip (10-core CPU, 16-core GPU) and 32 GB of unified memory. The operating system used was macOS Ventura 13.5. The implementation was written in Python 3.10. Key software libraries and frameworks include NumPy 1.24.2, SciPy 1.10.1, scikit-learn 1.2.2, and PyTorch 2.0.1 (CPU backend). Main experiments were run locally without the use of additional distributed or cloud-based resources.

\section{Related Work}
A handful of works have recently explored prediction sets in hierarchical label spaces, but HCC differs fundamentally in both goal and methodology. Prior methods, such as \citet{mortier2022set,mortier2025conformal,pmlr-v300-mortier26a}, aim to capture regions of high probability mass and are limited to trees, rather than our more general focus on coverage guarantees for DAGs. \citet{angelopoulosconformal} focuses on bounding the hierarchical distance rather than achieving leaf coverage, and their approach restricts the inference strictly to ancestor nodes of the most likely label. In contrast, our approach can flexibly include both internal and leaf nodes based on a user-specified trade-off between prediction set size and specificity. \citet{zhang2024conformal} adopt a learn-then-test framework called Conformal Structured Prediction to achieve a similar goal to ours, and provides coverage guarantees. Their method requires a user-specified limit on the nominal prediction set size $M$, requires a set of user-defined thresholds to test, and relies on an integer programming formulation that may not scale well with the taxonomy size. In contrast, our approach allows for explicitly trading off compactness with specificity and introduces a structured and substantially reduced search during calibration, which is further pruned of spurious solutions during inference in a scalable way. \citet{wieslander2020deep} combine deep learning with conformal prediction for the classification of tissue subregions in image classification. They assess the region-specific drug response with a hierarchical identification of these regions and measure confidence in subregion detection. However, this approach is specific to the problem of region of interest detection in the visual domain and does not use an existing class hierarchy.
Several works aim to improve the coverage for label spaces with an overlaid grouping function, these works do not involve hierarchical relationships we focus on here~\cite{bairaktari2025kandinsky,fargion2025enhancing,vovk2012conditional,ding2023class}.

In the context of multilabel classification problems, conformal prediction approaches include Binary Relevance Multi-Label Conformal Predictor (BR-MLCP) \citep{lambrou2016binary}, the instance-based method \citep{wang2014reliable}, and Power Set Multi-Label Conformal Predictor (PS-MLCP1) \citep{papadopoulos2014cross}. PS-MLCP1 has been applied in text classification tasks \citep{paisios2019deep}, and an extension of PS-MLCP1  effectively manages large labels  \citep{maltoudoglou2022well}.
BR-MLCP uses a one-against-all approach to break down a multilabel dataset into multiple single-label binary classification tasks. However, this approach fails to consider the dependencies between labels. On the other hand, PS-MLCP1 takes label dependencies into account but fails to address missing label information and faces computational issues when dealing with large labels. 
\citet{cauchois2021knowing} established a tree-structured technique to consider the interaction between the labels.
\citet{tyagi2023multi} proposed a tree-based method for multilabel classification problems using conformal prediction and multiple testing. 



\end{document}